\documentclass{article}
\usepackage{ijcai25}

\usepackage{times}
\usepackage{soul}
\usepackage{url}
\usepackage[hidelinks]{hyperref}
\usepackage[utf8]{inputenc}
\usepackage[small]{caption}
\usepackage{graphicx}
\usepackage{amsmath}
\usepackage{amsthm}
\usepackage{bm}
\usepackage{booktabs}
\usepackage{algorithm}
\usepackage{algorithmic}
\usepackage[switch]{lineno}
\usepackage{amssymb}
\usepackage{mathtools}
\usepackage{wrapfig}
\usepackage{subcaption}
\usepackage{lipsum}

\usepackage{titlesec}
\usepackage{xcolor}

\theoremstyle{plain}
\newtheorem{theorem}{Theorem}[section]

\newtheorem{lemma}[theorem]{Lemma}
\newtheorem{corollary}[theorem]{Corollary}
\theoremstyle{definition}

\newtheorem{assumption}[theorem]{Assumption}
\theoremstyle{remark}

\newcommand{\argmin}{\mathop{\rm argmin}\limits}

\newcommand{\bref}[1]{(\ref{#1})}
\newcommand{\set}[1]{\mathcal{#1}}
\newcommand{\balpha}{\boldsymbol{\alpha}}
\newcommand{\bbeta}{\boldsymbol{\beta}}
\newcommand{\bbalpha}{\bar{\boldsymbol{\alpha}}}
\newcommand{\bbbeta}{\bar{\boldsymbol{\beta}}}
\newcommand{\bbetad}{\boldsymbol{\beta}^{\dagger}}
\newcommand{\bbbetad}{\bar{\boldsymbol{\beta}}^{\dagger}}
\newcommand{\hatbbetad}{\hat{\boldsymbol{\beta}}^{\dagger}}
\newcommand{\bA}{\boldsymbol{A}}
\newcommand{\bB}{\boldsymbol{B}}
\newcommand{\bC}{\boldsymbol{C}}
\newcommand{\bD}{\boldsymbol{D}}
\newcommand{\bx}{\boldsymbol{x}}
\newcommand{\bxd}{\boldsymbol{x}^{\dagger}}

\newcommand{\by}{\boldsymbol{y}}

\newcommand{\bz}{\boldsymbol{z}}

\newcommand{\bI}{\boldsymbol{I}}

\newcommand{\tbyin}{\tilde{\boldsymbol{y}}^{\mathrm{H}}}

\newcommand{\tbyfe}{\tilde{\boldsymbol{y}}^{\mathrm{L}}}
\newcommand{\bZin}{\boldsymbol{Z}^{\mathrm{H}}}
\newcommand{\bZfe}{\boldsymbol{Z}^{\mathrm{L}}}

\newcommand{\bWin}{\boldsymbol{W}^{\mathrm{H}}}
\newcommand{\bWfe}{\boldsymbol{W}^{\mathrm{L}}}

\newcommand{\Lin}{\mathcal{L}_\mathrm{H}}
\newcommand{\Lfe}{\mathcal{L}_\mathrm{L}}
\newcommand{\Omegain}{\Omega_{\mathrm{H}}}
\newcommand{\Omegafe}{\Omega_{\mathrm{L}}}
\newcommand{\bv}{\boldsymbol{v}}

\newcommand{\bh}{\boldsymbol{h}}
\newcommand{\bone}{\boldsymbol{1}}
\newcommand{\bM}{\boldsymbol{M}}

\newcommand{\tbX}{\tilde{\boldsymbol{X}}}
\newcommand{\phiin}{\phi_{\mathrm{H}}}
\newcommand{\phife}{\phi_{\mathrm{L}}}
\newcommand{\bzero}{\boldsymbol{0}}

\newcommand{\tyin}{\tilde{y}^{\mathrm{H}}}

\newcommand{\tyfe}{\tilde{y}^{\mathrm{L}}}
\newcommand{\Dd}{D^{\dagger}}
\newcommand{\concat}{\mathtt{concat}}

\title{Explaining Black-box Model Predictions via Two-level Nested Feature Attributions with Consistency Property}

\author{
Yuya Yoshikawa$^1$\thanks{This manuscript is an extended version of our paper accepted at IJCAI~2025, with detailed proofs and additional experimental results.}\and
Masanari Kimura$^2$\and
Ryotaro Shimizu$^{3}$\And
Yuki Saito$^{3}$\\
\affiliations
$^1$STAIR Lab, Chiba Institute of Technology\\
$^2$School of Mathematics and Statistics, The University of Melbourne\\
$^3$ZOZO Research\\
\emails
yoshikawa@stair.center,
m.kimura@unimelb.edu.au,
ryotaro.shimizu@zozo.com,
yuki.saito@zozo.com
}

\begin{document}

\maketitle

\begin{abstract}
Techniques that explain the predictions of black-box machine learning models are crucial to make the models transparent, thereby increasing trust in AI systems. 
The input features to the models often have a nested structure that consists of high- and low-level features, and each high-level feature is decomposed into multiple low-level features.
For such inputs, both high-level feature attributions (HiFAs) and low-level feature attributions (LoFAs) are important for better understanding the model's decision.
In this paper, we propose a model-agnostic local explanation method that effectively exploits the nested structure of the input to estimate the two-level feature attributions simultaneously.
A key idea of the proposed method is to introduce the consistency property that should exist between the HiFAs and LoFAs, thereby bridging the separate optimization problems for estimating them.
Thanks to this consistency property, the proposed method can produce HiFAs and LoFAs that are both faithful to the black-box models and consistent with each other, using a smaller number of queries to the models.
In experiments on image classification in multiple instance learning and text classification using language models, we demonstrate that the HiFAs and LoFAs estimated by the proposed method are accurate, faithful to the behaviors of the black-box models, and provide consistent explanations.
\end{abstract}

\section{Introduction}\label{sec:intro}
The rapid increase in size and complexity of machine learning (ML) models has led to a growing concern about their {\it black-box} nature. 
Models provided as cloud services are literal black boxes, as users have no access to the models themselves or the training data used. 
This opacity raises numerous concerns, including issues of trust, accountability, and transparency. 
Consequently, techniques for explaining the predictions made by such black-box models have been attracting significant attention~\cite{Danilevsky2020-sy,Dosilovic2018-uh,Saeed2023-pn}. 

Various {\it model-agnostic} local explanation methods have been proposed to explain the predictions of black-box models.
Representative methods include, for example, local interpretable model-agnostic explanation (LIME)~\cite{Ribeiro2016-nj} and kernel Shapley additive explanations (Kernel SHAP)~\cite{Lundberg2017-ii}, which estimate the feature attributions for the individual prediction by approximating the model's behavior with local linear surrogate models around the input.

In LIME and Kernel SHAP, the input to the model is generally assumed to be a flat structure, where the input features are treated as independent variables.
In many realistic tasks for various domains, such as image, text, geographic, e-commerce, and social network data, however, the input features have a nested structure that consists of high- and low-level features, and each high-level feature is decomposed into multiple low-level features.
A typical task with such nested features is multiple instance learning (MIL)~\cite{Ilse2018-be} where the model is formulated as a set function~\cite{Kimura2024-wq}.
In MIL, the input is a set of instances, the high-level feature is an instance in the set, and the low-level features represent the features of the instance.
In addition, even if the input is not represented in a nested structure when it is fed into the model, it may be more natural to interpret it with the nested structure.
For example, although a text input is usually represented as a sequence of words, it is natural to interpret it as having high-level features such as phrases, sentences, and paragraphs.

\begin{figure*}[t]
\centering
\includegraphics[width=0.8\linewidth]{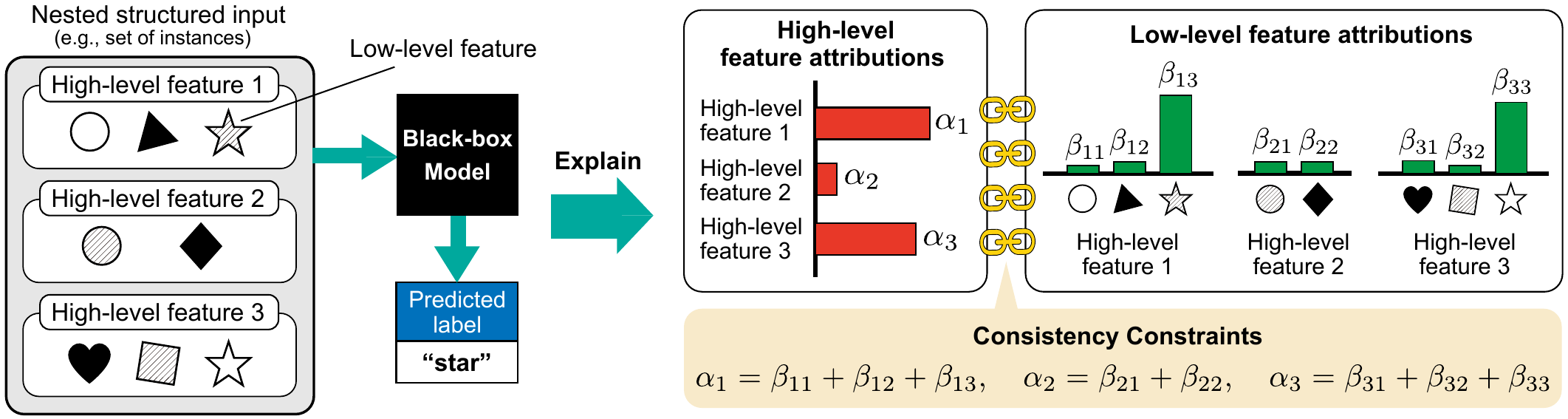}
\caption{Example of the black-box model prediction for a nested structured input and its corresponding high- and low-level feature attributions estimated by the proposed method with consistency constraints.
Objects in each high-level feature represent the low-level features.
}
\label{fig:intro:example}
\end{figure*}

The two-level features enable us to understand the model predictions with two types of feature attributions at different levels of granularity, which we name {\it high-level feature attributions (HiFAs)} and {\it low-level feature attributions (LoFAs)}, respectively.
Figure~\ref{fig:intro:example} shows an example of the prediction for a nested-structured input and its corresponding HiFAs and LoFAs.
The HiFAs represent how much each of the high-level features contributes to the prediction.
These are also referred to as {\it instance attributions} in the MIL literature~\cite{Early2022-oc,Javed2022-sm}, which are used to reveal which instances strongly affected the model's decision.
On the other hand, the LoFAs represent how much each of the low-level features contributes to the prediction, providing a finer-grained explanation of how the components of the instances affected the prediction. 
Both the HiFAs and LoFAs are important for understanding the model's decision.
{\it However, existing studies have focused on estimating either-level attributions, and no study has addressed estimating the HiFAs and LoFAs simultaneously.}

For the estimation of the HiFAs and LoFAs, two naive approaches can be applied.
One is to estimate the HiFAs and LoFAs separately by applying existing model-agnostic local explanation methods to the high- and low-level features, respectively.
The other is to estimate the LoFAs first, as in the former approach, and then estimate the HiFAs by aggregating the LoFAs.
However, these approaches have two limitations in terms of leveraging the nested structure of the input.
First, even though the queries to the black-box model are often limited in practice due to the computational time and request costs, the input structure is not utilized to reduce the number of queries in the estimation.
Second, the former approach can produce mutually inconsistent HiFAs and LoFAs. 
For example, the high-level feature identified as most influential (highest HiFA) might not coincide with the high-level feature that contains the most influential low-level feature (highest LoFA).

To address these issues, we propose a model-agnostic local explanation method that effectively exploits the nested structure of the input to estimate the HiFAs and LoFAs simultaneously.
A key idea of the proposed method is to introduce the consistency property that should exist between the HiFAs and LoFAs, thereby bridging the separate optimization problems for them.
We solve a joint optimization problem to estimate the HiFAs and LoFAs simultaneously with the consistency constraints depicted in Figure~\ref{fig:intro:example} based on the alternating direction method of multipliers (ADMM)~\cite{Boyd_undated-bc}.
The algorithm is a general framework that can also introduce various types of regularizations and constraints for the HiFAs and LoFAs, such as the $\ell_1$ and $\ell_2$ regularizations and non-negative constraints, which lead to the ease of interpretability for humans.

In experiments, we quantitatively and qualitatively assess the HiFAs and LoFAs estimated by the proposed method on image classification in the MIL setting and text classification using language models.
The experimental results show that the HiFAs and LoFAs estimated by the proposed method 1) satisfy the consistency property, 2) are faithful explanations to the black-box models even when the number of queries to the model is small, 3) can accurately guess the ground-truth positive instances and their features in the MIL task, and 4) are reasonable explanations visually.

The contributions of this work are summarized as follows:
{
\setlength{\leftmargini}{12pt}         %
\begin{itemize}
	\setlength{\itemsep}{5pt}      %
	\setlength{\parskip}{0pt}      %
  \item This study is the first to propose a model-agnostic local explanation method to estimate the two-level nested feature attributions simultaneously, which satisfies the consistency property between them.
  \item To justify the behavior of the proposed method, we provide theoretical analysis in terms of the existence and uniqueness of the solution and the convergence of the optimization algorithm. 
  \item In the experiments on practical tasks, we demonstrate that the proposed method can produce accurate, faithful, and consistent two-level feature attributions with a smaller number of queries to the black-box models.
\end{itemize}
}

\section{Related Work}\label{sec:RW}

Numerous methods for explaining the individual predictions of black-box models have been proposed in the literature~\cite{Ribeiro2016-nj,Lundberg2017-ii,Ribeiro2018-ws,Petsiuk2018-uj,Plumb2018-ac}.
A versatile approach is to explain feature attributions estimated by approximating the model predictions with surrogate models around the input, such as LIME~\cite{Ribeiro2016-nj} and Kernel SHAP~\cite{Lundberg2017-ii}.
The proposed method is in line with this type of approach.

One example of a nested input structure is set data, where a set of multiple instances is treated as a single input.
Set data appears in various ML applications, such as point cloud classification~\cite{Guo2021-vp}, medical image analysis~\cite{Cheplygina2018-wj}, and group recommendation~\cite{Dara2020-ml}, and the explainability in those applications has also been studied in the literature~\cite{Tan2022-ki,Van_der_Velden2022-mk}.
Unlike our work, most such studies focus only on estimating instance attributions corresponding to those of high-level features.
For example, Early et al. proposed to estimate instance attributions by learning surrogate models with MIL-suitable kernel functions~\cite{Early2022-oc}.

Several studies have addressed estimating feature attributions effectively by leveraging group information of input features.
In the natural language processing literature, some studies estimated sentence- and phrase-level feature attributions by grouping words in the same sentence and phrase together and regarding them as a single feature~\cite{Zafar2021-aq,Mosca2022-ki}.
In addition, Rychener et al. showed that word-level feature attributions can be improved by generating perturbations at a sentence level, mitigating out-of-distribution issues for the model and the challenges of a high-dimensional search space~\cite{Rychener2020-gu}.
In the official SHAP library~\cite{noauthor_undated-ca}, by grouping input features by hierarchical clustering in advance and generating perturbations at the group level, one can reduce the number of queries to the model.

\section{Proposed Method}\label{sec:proposed}
\subsection{Two-level Nested Feature Attributions with Surrogate Models}

The model $f$ to be explained is a trained black-box model that takes an arbitrary input, such as tabular, image or text, $\bx \in \set{X}$, and outputs a prediction $\by = f(\bx) \in [0,1]^{C}$ where $\set{X}$ is the input space, and $C$ is the number of classes.
The input $\bx$ consists of two-level nested features, referred to as high-level and low-level features, with each high-level feature being decomposed into multiple low-level features.
In particular, the input $\bx$ is represented as a set or sequence of $J$ high-level features, i.e., $\bx = \{ \bx_j \}_{j=1}^{J}$ where $\bx_j \in \mathbb{R}^{D_j}$ is a $D_j$-dimensional low-level feature vector representing the $j$-th high-level feature.
One example of such input appears in image classification under the MIL setting. 
In this setting, the input is a bag of images, the high-level feature is an image in the bag, and the low-level features correspond to super-pixels in the image.
Another example appears in a document classification where the input is a sequence of sentences, the high-level feature is a sentence in the sequence, and the low-level features correspond to the words in the sentence.

We consider estimating the high-level feature attributions (HiFAs) and low-level feature attributions (LoFAs) that explain the prediction of the black-box model $f$ for the input $\bx$ using surrogate models as with LIME and Kernel SHAP.
The HiFAs and LoFAs represent how much the high- and low-level features in the input contribute to the prediction, respectively.
In the aforementioned MIL setting, the HiFAs represent how much images in the input bag contribute to the prediction, which is also referred to as {\it instance attributions} in the literature, and the LoFAs represent how much the super-pixels in the images contribute to the prediction.  %
To estimate the HiFAs and LoFAs, we introduce two-level local linear surrogate models for high-level and low-level features, $e^{\mathrm{H}}$ and $e^{\mathrm{L}}$, that mimic the behaviors of the black-box model $f$ around the input $\bx$, as follows:
\begin{equation}
e^{\mathrm{H}}(\bz^{\mathrm{H}}; \balpha) = \sum_{j=1}^J \alpha_j z^{\mathrm{H}}_j,\quad\quad
e^{\mathrm{L}}(\bz^{\mathrm{L}}; \bbeta) = \sum_{j=1}^J \sum_{d=1}^{D_j} \beta_{jd} z^{\mathrm{L}}_{jd},
\label{eq:e}
\end{equation}
where $\bz^{\mathrm{H}} \in \{0,1\}^J$ and $\bz^{\mathrm{L}} = \{ \bz^{\mathrm{L}}_j \}_{j=1}^{J}$ with $\bz^{\mathrm{L}}_j \in \{0,1\}^{D_j}$ are simplified inputs associated with the input $\bx$, which are used to indicate the presence or absence of the high- and low-level features in $\bx$, respectively; $\balpha \in \mathbb{R}^{J}$ and $\bbeta = \{ \bbeta_j \}_{j=1}^{J}$ with $\bbeta_j \in \mathbb{R}^{D_j}$ are the learnable coefficients of these surrogate models, and after learning, they will be the HiFAs and LoFAs themselves, respectively.
For ease of computation, we define the concatenation of $\bbeta$ and $\bz^{\mathrm{L}}$ over the high-level features as $\bbetad = \concat(\bbeta_1, \bbeta_2, \cdots, \bbeta_J) \in \mathbb{R}^{\Dd}$ and $\bz^{\mathrm{L}\dagger} = \concat(\bz^{\mathrm{L}}_1, \bz^{\mathrm{L}}_2, \cdots, \bz^{\mathrm{L}}_J) \in \{0,1\}^{\Dd}$, where $\Dd = \sum_{j=1}^{J} D_j$.

The surrogate models are learned with the predictions of the black-box model $f$ for perturbations around the input $\bx$.
The perturbations are generated by sampling the simplified inputs $\bz^{\mathrm{H}}$ and $\bz^{\mathrm{L}\dagger}$ from binary uniform distributions and then constructing masked inputs $\phi_{\bx}^{\mathrm{H}}(\bz^{\mathrm{H}}), \phi_{\bx}^{\mathrm{L}}(\bz^{\mathrm{L}\dagger}) \in \set{X}$ depending on the simplified inputs, respectively.
Here, $\phi_{\bx}^{\mathrm{H}}$ and $\phi_{\bx}^{\mathrm{L}}$ are mask functions that replace the input $\bx$'s dimensions associated with the dimensions being zero in the simplified inputs $\bz^{\mathrm{H}}$ and $\bz^{\mathrm{L}\dagger}$ with uninformative values, such as zero, respectively.
Let $\bZin \in \{0,1\}^{N_{\rm{H}} \times J}$ and $\bZfe \in \{0,1\}^{N_{\rm{L}} \times \Dd}$ be the matrices whose rows are the generated simplified inputs for the high- and low-level features, respectively, where $N_{\rm{H}}$ and $N_{\rm{L}}$ are the numbers of perturbations used to estimate the HiFAs and LoFAs, respectively.
Also, let $\tbyin = [ \tyin_1, \tyin_2, \cdots, \tyin_{N_{\rm{H}}} ]^{\top} \in \mathbb{R}^{N_{\rm{H}} \times C}$ and $\tbyfe = [ \tyfe_1, \tyfe_2, \cdots, \tyfe_{N_{\rm{L}}} ]^{\top} \in \mathbb{R}^{N_{\rm{L}} \times C}$ be the predictions of the black-box model for the perturbations where $\tyin_n = f(\phi_{\bx}^{\mathrm{H}}(\bZin_n))$ and $\tyfe_n = f(\phi_{\bx}^{\mathrm{L}}(\bZfe_n))$.

The parameters of the surrogate models, i.e., the HiFAs $\hat{\balpha}$ and LoFAs $\hat{\bbetad}$, can be estimated by solving the following weighted least squares separately:
\begin{align}
\label{eq:proposed:HiFA}
&\hat{\balpha}
= \argmin_{\balpha} \Lin(\balpha) + \lambda_{\rm{H}} \Omega_{\rm{H}}(\balpha)\\
&\quad\text{where}\quad
\Lin(\balpha) = \frac{1}{2}(\tbyin - \bZin \balpha)^{\top} \bWin (\tbyin - \bZin \balpha),\nonumber
\\
\label{eq:proposed:LoFA}
&\hat{\bbetad}
= \argmin_{\bbetad} \Lfe(\bbetad) + \lambda_{\rm{L}} \Omega_{\rm{L}}(\bbetad)\\
&\quad\text{where}\quad
\Lfe(\bbetad) = \frac{1}{2}(\tbyfe - \bZfe \bbetad)^{\top} \bWfe (\tbyfe - \bZfe \bbetad),\nonumber
\end{align}
where $\bWin \in \mathbb{R}^{N_{\rm{H}} \times N_{\rm{H}}}$ and $\bWfe \in \mathbb{R}^{N_{\rm{L}} \times N_{\rm{L}}}$ are the diagonal matrices whose $n$th diagonal elements represent the sample weights for the $n$th perturbation; $\Omega_{\rm{H}}$ and $\Omega_{\rm{L}}$ are the regularizers for the HiFAs and LoFAs, respectively; $\lambda_{\rm{H}} \geq 0$ and $\lambda_{\rm{L}} \geq 0$ are the regularization strengths.

\subsection{Optimization with Consistency Constraints}

Although the HiFAs and LoFAs provide different levels of explanations, these explanations for the same black-box model should be consistent with each other.
Owing to the linearity of the surrogate models and the fact that each high-level feature can be decomposed into low-level features, the following property is expected to hold:
\theoremstyle{definition}
\newtheorem{dfn}{Property}
\begin{dfn}[Consistency between two-level feature attributions]

\begin{equation}
\alpha_j = \sum_{d=1}^{D_j} \beta_{jd} \quad (\forall j \in [J]).
\label{eq:proposed:consistency}
\end{equation}
In other words, the HiFA for each high-level feature is equal to the sum of the LoFAs of its constituent low-level features.
The two surrogate models defined in~\bref{eq:e} that satisfy the consistency property behave equivalently for simplified inputs $\bz^{\mathrm{H}}$ and $\bz^{\mathrm{L}}$ under the condition if $z^{\mathrm{H}}_j = 0$ then $\bz^{\mathrm{L}}_j = \bzero_{D_j}$, and if $z^{\mathrm{H}}_j = 1$ then $\bz^{\mathrm{L}}_j = \bone_{D_j}$ where $\bzero_{D_j}$ and $\bone_{D_j}$ are the $D_j$-dimensional zero and one vectors, respectively.
\end{dfn}

The consistency property is essential for providing consistent and convincing explanations to humans. 
However, in practice, it is often not satisfied for two reasons.
First, the number of perturbations may be insufficient to accurately estimate the feature attributions because the number of queries to the model $f$ is often limited due to computational time and request costs.
Second, in the predictions for the perturbations, the model $f$’s behavior can differ depending on whether the high-level features or the low-level features are masked out because of {\it missingness bias}~\cite{Jain2022-oe}.
To overcome these problems, the proposed method estimates the HiFAs and LoFAs simultaneously by solving the following optimization with consistency constraints:
\begin{align}
\hat{\balpha},\hatbbetad &= \argmin_{\balpha,\bbetad} \Lin(\balpha) + \Lfe(\bbetad) 
+ \lambda_{\rm{H}} \Omega_{\rm{H}}(\balpha) + \lambda_{\rm{L}}\Omega_{\rm{L}}(\bbetad)\nonumber\\
&\quad\mathrm{s.t.}\quad \alpha_j = \sum_{d=1}^{D_j} \beta_{jd} \quad (\forall j \in [J]).
\label{eq:proposed:objective1}
\end{align}
The consistency constraints bridge the gap between the two surrogate models, forcing them to behave equivalently.
This helps compensate the insufficiency of the queries to the model and mitigate the negative effects of the missingness bias on the estimation of the HiFAs and LoFAs.

We solve the optimization problem based on the alternating direction method of multipliers (ADMM)~\cite{Boyd_undated-bc}.
The detailed derivation of the optimization algorithm is provided in Appendix~\ref{sec:appendix:optimization}.
An advantage of employing the ADMM is that, despite the interdependence of $\balpha$ and $\bbetad$ caused by the consistency constraints, they can be estimated independently as in \bref{eq:proposed:HiFA} and \bref{eq:proposed:LoFA}.
In addition, the solution has another advantage: it allows the inclusion of various types of regularizations and constraints for $\balpha$ and $\bbeta$, such as sparsity regularization and non-negative constraints in $\Omega_{\rm{H}}$ and $\Omega_{\rm{L}}$.
The optimization algorithm is provided in Algorithm~\ref{alg:appendix:optimization} in Appendix~\ref{sec:appendix:optimization}.
The computational time complexity of the algorithm is discussed in Appendix~\ref{sec:appendix:complexity}. %

\section{Auxiliary Theoretical Analysis}
\label{sec:auxiliary_theoretical_analysis}
In this section, we provide theoretical analysis for the proposed method, referred to as {\it Consistent Two-level Feature Attribution (C2FA)}, to justify its behavior.
See Appendix~\ref{sec:appendix:proofs} for detailed proofs of all statements in this section.  %
In this section, we impose the following assumptions.
\begin{assumption}[Convexity]
    \label{asm:convexity}
    $\Lin(\cdot)$, $\Lfe(\cdot)$, $\Omegain(\cdot)$, $\Omegafe(\cdot)$ are convex in their respective parameters.
    Specifically, each loss term is weighted least squares loss, and each regularizer is a convex function.
\end{assumption}
\begin{assumption}[Strict Feasibility]
    \label{asm:strict_feasibility}
    There exists $(\balpha^\circ, \bbeta^\circ)$ such that $\alpha^\circ_j = \sum^{D_j}_{d=1}\beta^\circ_{j,d}$ for all $j$, ensuring the constraint set is non-empty.
\end{assumption}
\begin{assumption}[Non-degeneracy]
    \label{asm:non_degeneracy}
    The design matrices underlying $\bz^H$ and $\bz^L$ have full column rank in typical regression settings.
    This ensures identifiability of $\balpha$ and $\bbeta$.
\end{assumption}
\begin{assumption}[Bounded Outputs]
    \label{asm:bounded_outputs}
    Black-box outputs $\{f(x_i)\}^N_{i=1}$ lie in a bounded set, or at least have finite second moments, ensuring the objective is well-defined and finite.
\end{assumption}
\begin{assumption}[Parameter Domains]
    \label{asm:parameter_domains}
    $\balpha$ and $\bbeta$ lie in closed, convex domains $\Gamma_{\balpha}$ and $\Gamma_{\bbeta}$.
\end{assumption}
Define the feasible set
\begin{equation*}
    \Gamma^\circ \coloneqq \left\{(\balpha, \bbeta) \mid \alpha_j = \sum^{D_j}_{d=1}\beta_{j,d},\ \forall j, \balpha \in \Gamma_{\balpha}, \bbeta \in \Gamma_{\bbeta} \right\},
\end{equation*}
and define $\mathcal{J}(\balpha, \bbeta) \coloneqq \Lin(\balpha) + \Lfe(\bbeta) + \lambda_H \Omegain(\balpha) + \lambda_L \Omegafe(\bbeta)$.
Hence the optimization problem is $\min_{(\balpha, \bbeta) \in \Gamma^\circ}\mathcal{J}(\balpha, \bbeta)$.
We develop existence and uniqueness of solutions under mild conditions, as well as the precise satisfaction of $\alpha_j = \sum^{D_j}_{d=1}\beta_{j,d}$ at optimality.
We first establish that a global minimizer exists. Under additional strict convexity conditions, we also show uniqueness
\begin{lemma}[Existence of a Global Minimizer]
    \label{lem:existence_of_a_global_minimizer}
    Under assumptions, there exists at least one global minimizer $(\balpha^*, \bbeta^*)$ of C2FA optimization.
\end{lemma}
\begin{lemma}[Strong Convexity and Uniqueness]
    \label{lem:strong_convexity_and_uniqueness}
    Suppose that the high-level and low-level losses $\Lin$ and $\Lfe$ are strictly convex quadratic forms in $\balpha$ and $\bbeta$, respectively (e.g., weighted least-squares with full column rank), and $\Omegain, \Omegafe$ are either strictly convex or affine so as not to destroy strict convexity of the overall objective.
    Then any global minimizer $(\balpha^*, \bbeta^*)$ is unique.
\end{lemma}
\begin{theorem}
    \label{thm:strictly_reduce_objective}
    Let $\tilde{\balpha}$ be an unconstrained minimizer of $\Lin(\balpha) + \lambda_H \Omegain(\balpha)$, and let $\tilde{\balpha}$ be an unconstrained minimizer of $\Lfe(\bbeta) + \lambda_L \Omegafe(\bbeta)$.
    Then, if $(\tilde{\balpha}, \tilde{\bbeta})$ do not already satisfy $\tilde{\alpha}_j = \sum_d\tilde{\beta}_{j,d}$ for each $j$, any global minimizer $(\balpha^*, \bbeta^*)$ of the C2FA optimization achieves strictly smaller objective value: $\mathcal{J}(\balpha^*, \bbeta^*) < \mathcal{J}(\tilde{\balpha}, \tilde{\bbeta})$.
\end{theorem}

Next, we provide the convergence analysis based on the concentration inequality.
\begin{theorem}[High-Probability Convergence of C2FA]
    \label{thm:hp-convergence-c2fa}
    Let $\{(\mathbf{z}^{\mathrm{H}}_n, Y^{\mathrm{H}}_n)\}_{n=1}^{N_{\mathrm{H}}}$ 
    and 
    $\{(\mathbf{z}^{\mathrm{L}}_n, Y^{\mathrm{L}}_n)\}_{n=1}^{N_{\mathrm{L}}}$
    be i.i.d.\ samples drawn from some distribution of mask vectors (for high-level and low-level features) and black-box outputs, with sub-Gaussian or bounded noise. 
    Form the matrices 
    $\mathbf{Z}^{\mathrm{H}} \in \mathbb{R}^{N_{\mathrm{H}}\times J}$ 
    and 
    $\mathbf{Z}^{\mathrm{L}} \in \mathbb{R}^{N_{\mathrm{L}}\times \Dd}$ 
    by stacking $\mathbf{z}^{\mathrm{H}}_n,\mathbf{z}^{\mathrm{L}}_n$, 
    and similarly stack $Y^{\mathrm{H}},Y^{\mathrm{L}}$.
    Suppose the following:
    \begin{itemize}
    \item[(i)] The random design satisfies a restricted eigenvalue property on $\mathbf{Z}^{\mathrm{H}}$ and $\mathbf{Z}^{\mathrm{L}}$ with high probability.
    \item[(ii)] The losses $\Lin$ and $\Lfe$ are given by weighted least squares, 
    and the regularizers $\Omega_{\mathrm{H}}, \Omega_{\mathrm{L}}$ are convex. 
    \item[(iii)] $\alpha_j=\sum_{d=1}^{D_j}\beta_{jd}$ for each $j=1,\ldots,J$.
    \item[(iv)] Let $(\balpha^\star,\bbeta^\star)$ be the minimizer of the expected version of the same objective under that constraint, i.e.\ the best consistent linear approximation of $f$ in expectation.
    \end{itemize}
    Then, for any $\delta\in(0,1)$, with probability at least $1-\delta$, the solution 
    $(\widehat{\balpha},\widehat{\bbeta})$ to the finite-sample C2FA problem satisfies
    \begin{align*}
    \bigl\|\bigl(\widehat{\balpha}, \widehat{\bbeta}\bigr) 
      \;-\;
           \bigl(\balpha^\star,\bbeta^\star\bigr)\bigr\|_2
    \;\;\le\;\;
    C\,
    \sqrt{\frac{(J+\Dd)\;+\;\log\!\bigl(\tfrac{1}{\delta}\bigr)}{\,N_{\mathrm{H}} + N_{\mathrm{L}}\,}},
    \end{align*}
    where $C>0$ is a constant depending on sub-Gaussian parameters, the restricted eigenvalue lower bound, and the regularization strengths.
\end{theorem}

\begin{corollary}[Uniform Approximation Guarantee]
    \label{cor:uniform-approx}
    Assume the same setting as Theorem~\ref{thm:hp-convergence-c2fa} and let 
    $e^{\mathrm{H}}(\mathbf{z}^{\mathrm{H}}) = (\mathbf{z}^{\mathrm{H}})^\top \widehat{\balpha}$ and $e^{\mathrm{L}}(\mathbf{z}^{\mathrm{L}}) = (\mathbf{z}^{\mathrm{L}})^\top \widehat{\bbeta}$  be the final C2FA surrogate models.  
    Further suppose that $\mathbf{z}^{\mathrm{H}}$ and $\mathbf{z}^{\mathrm{L}}$ take values in some bounded set $\mathcal{Z}^{\mathrm{H}} \subseteq \mathbb{R}^J$ and $\mathcal{Z}^{\mathrm{L}} \subseteq \mathbb{R}^{\Dd}$ respectively. 
    Then with probability at least $1-\delta$,
    \begin{align*}
        &\sup_{\mathbf{z}^{\mathrm{H}}\in\mathcal{Z}^{\mathrm{H}}} \Bigl|\,f\bigl(\phi^{\mathrm{H}}_{\mathbf{x}}(\mathbf{z}^{\mathrm{H}})\bigr) - e^{\mathrm{H}}(\mathbf{z}^{\mathrm{H}})\Bigr| + \sup_{\mathbf{z}^{\mathrm{L}}\in\mathcal{Z}^{\mathrm{L}}} \Bigl|\,f\bigl(\phi^{\mathrm{L}}_{\mathbf{x}}(\mathbf{z}^{\mathrm{L}})\bigr) - e^{\mathrm{L}}(\mathbf{z}^{\mathrm{L}})\Bigr| \\
        &\quad\quad\quad\quad\quad\quad \leq  B_H\cdot\widetilde{C} \sqrt{\frac{(J+\Dd)\;+\;\log\!\bigl(\tfrac{1}{\delta}\bigr)}{\,N_{\mathrm{H}} + N_{\mathrm{L}}\,}} + \epsilon^*,
    \end{align*}
    for some constant $\widetilde{C}, B_H, \epsilon^*$ depending on sub-Gaussian parameters, bounding sets $\mathcal{Z}^{\mathrm{H}},\mathcal{Z}^{\mathrm{L}}$, and the regularization strengths.
\end{corollary}

\section{Experiments}\label{sec:experiment}
We conducted experiments on two tasks of image and text domains to evaluate the effectiveness of the proposed method, referred to as C2FA, implemented with Algorithm~\ref{alg:appendix:optimization} in Appendix~\ref{sec:appendix:optimization}.
Its hyperparameters are provided in Appendix~\ref{sec:appendix:image:implementation}.

\paragraph{Comparing Methods.}
We used five methods for comparison: LIME~\cite{Ribeiro2016-nj}, MILLI~\cite{Early2022-oc}, Bottom-Up LIME (BU-LIME), Top-Down LIME (TD-LIME), and Top-Down MILLI (TD-MILLI).
With LIME, we estimated the HiFAs and LoFAs separately by solving~\bref{eq:proposed:HiFA} and~\bref{eq:proposed:LoFA}, respectively, where we used the cosine kernel for the sample weights and $\ell_2$ regularization for $\Omega_{\rm{H}}$ and $\Omega_{\rm{L}}$.
MILLI is the state-of-the-art instance attribution method in the MIL setting, which was proposed for estimating only the HiFAs.
Therefore, we estimated the LoFAs in MILLI in the same way as in LIME.
With BU-LIME, we first estimated the LoFAs using LIME and then calculated the HiFAs of each high-level feature by summing the LoFAs associated with the high-level feature.
This method always satisfies the consistency property because the HiFAs are calculated from the LoFAs.
With TD-LIME and TD-MILLI, we first estimated the HiFAs using LIME and MILLI, respectively.
Then, for the $j$th high-level feature, we determined its LoFAs $\bbeta_j$ by sampling from a normal distribution with mean $\alpha_j$ (the $j$th HiFA) and the standard deviation $1 / D_j$.
Finally, we enforced the consistency property by adjusting one randomly selected low-level feature for each high-level feature $j$: we set that feature's LoFA to $\beta_{jd} = \alpha_j - \sum_{d' \in [D_j] \backslash \{d\}} \beta_{jd'}$, ensuring that $\sum_{d' \in [D_j]} \beta_{jd'} = \alpha_j$.

\subsection{Multiple Instance Image Classification}\label{sec:experiment:image}

\begin{figure*}[t]
\centering
\begin{subfigure}{0.25\textwidth}
  \centering
  \includegraphics[width=\textwidth]{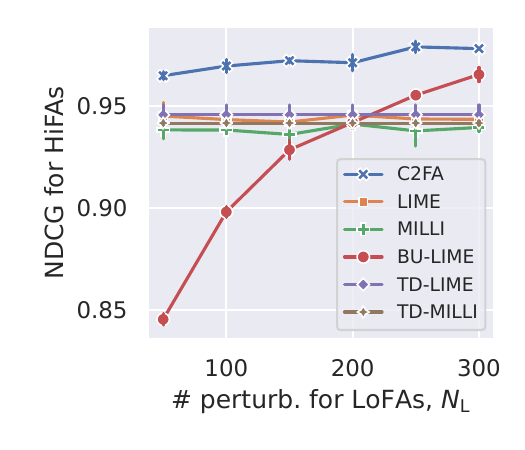}\\
  \includegraphics[width=\textwidth]{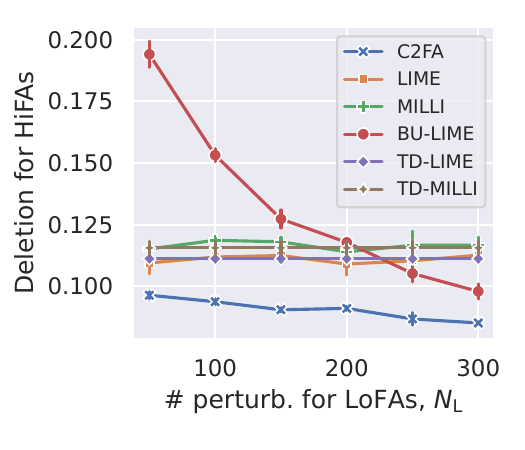}
  \caption{}
  \label{fig:experiment:vocbag:HiFA}
\end{subfigure}
\begin{subfigure}{0.25\textwidth}
  \centering
  \includegraphics[width=\textwidth]{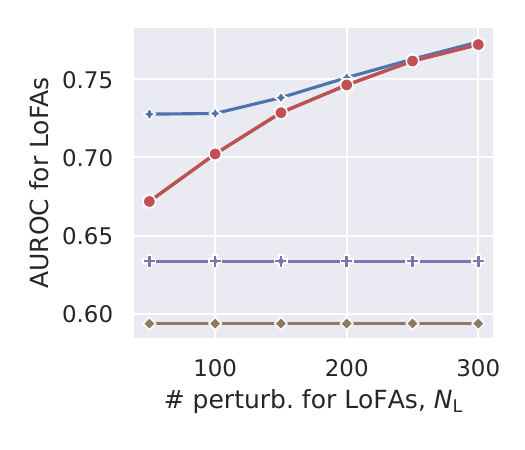}\\
  \includegraphics[width=\textwidth]{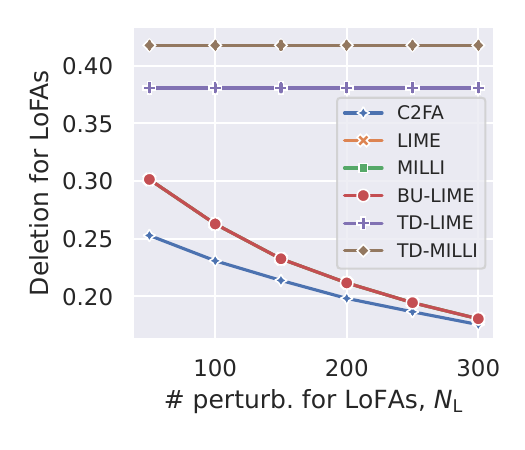}
  \caption{}
  \label{fig:experiment:vocbag:LoFA}
\end{subfigure}
\begin{subfigure}{0.25\textwidth}
  \centering
  \includegraphics[width=\textwidth]{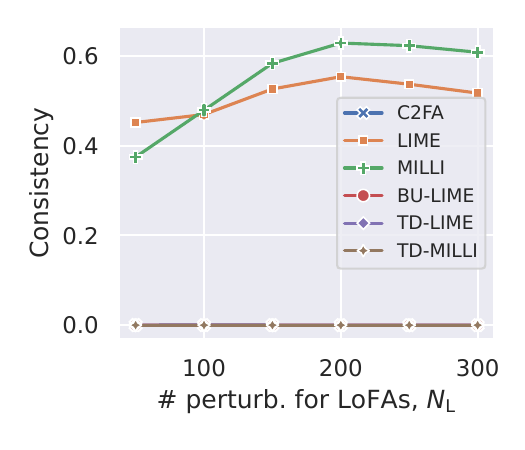}\\
  \includegraphics[width=\textwidth]{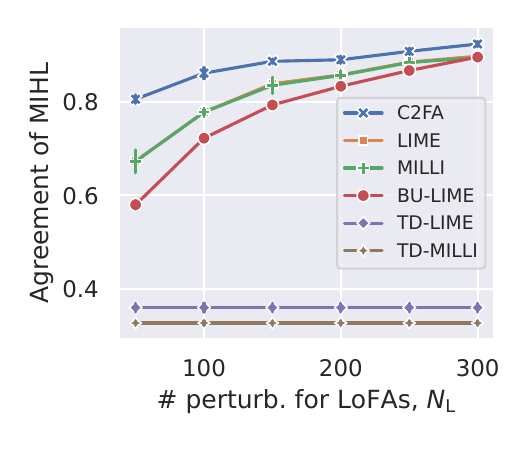}
  \caption{}
  \label{fig:experiment:vocbag:consistency}
\end{subfigure}
\caption{
  Quantitative evaluation on the image classification task.
  (a) NDCG (higher is better) and deletion scores (lower is better) of the estimated HiFAs.
  (b) AUROC (higher is better) and deletion scores (lower is better) of the estimated LoFAs.
  (c) Consistency scores (lower is better) and the agreement scores of MIHL (higher is better).
  The error bars represent the standard deviations of the scores over three runs with different random seeds.
}
\label{fig:experiment:vocbag}
\end{figure*}

\paragraph{Dataset.}
We constructed an MIL dataset from the Pascal VOC semantic segmentation dataset~\cite{Everingham15} with the ground-truth instance- and pixel-level labels.
Each sample (bag) contains three to five images, randomly selected from the Pascal VOC dataset.
Here, low-level features correspond to regions (super-pixels) of each image.
Each bag is labeled positive if at least one image in the bag is associated with the ``cat'' label; otherwise the bag is negative.
Also, each image pixel is labeled positive if it is associated with the ``cat'' label, and negative otherwise.
A detailed description of the dataset is provided in Appendix~\ref{sec:appendix:image:dataset}.

\paragraph{Black-box Model.}
We used a DeepSets permutation-invariant model~\cite{Zaheer2017-in} with ResNet-50~\cite{He2016-kb} as the black-box model $f$ to be explained.
We describe the implementation details of the model in Appendix~\ref{sec:appendix:image:implementation}.
Here, the test accuracy of the model was 0.945.

\paragraph{Quantitative Evaluation.}
We assessed the estimated HiFAs and LoFAs in terms of correctness, faithfulness, and consistency.
Correctness is evaluated using the ground-truth instance- and pixel-level labels.
Following the evaluation in the MIL study~\cite{Early2022-oc}, we evaluated the estimated HiFAs with normalized discounted cumulative gain (NDCG).
Additionally, by regarding the estimated LoFAs as the predictions of the pixel-level labels, we evaluated them with the area under ROC curve (AUROC) in a binary semantic segmentation setting.
In the faithfulness evaluation, we assessed whether the estimated HiFAs and LoFAs are faithful to the behaviors of the model $f$ based on insertion and deletion metrics~\cite{Petsiuk2018-uj}.
For consistency, we used the following two metrics.
The first one is the consistency between the estimated HiFAs and LoFAs, measured by $\| \balpha - \bM \bbetad \|^2$, which is the same quantity used as the consistency penalty in~\bref{eq:appendix:objective2}.
The second one is the agreement of the most important high- and low-level feature (MIHL).
We ran the evaluations three times with different random seeds and reported the average scores and their standard deviations.
The details of the evaluation metrics are provided in Appendix~\ref{sec:appendix:image:evaluation}.

\begin{figure}[!h]
\scriptsize
\centering
Input (bag of images)\\
\includegraphics[width=0.85\columnwidth]{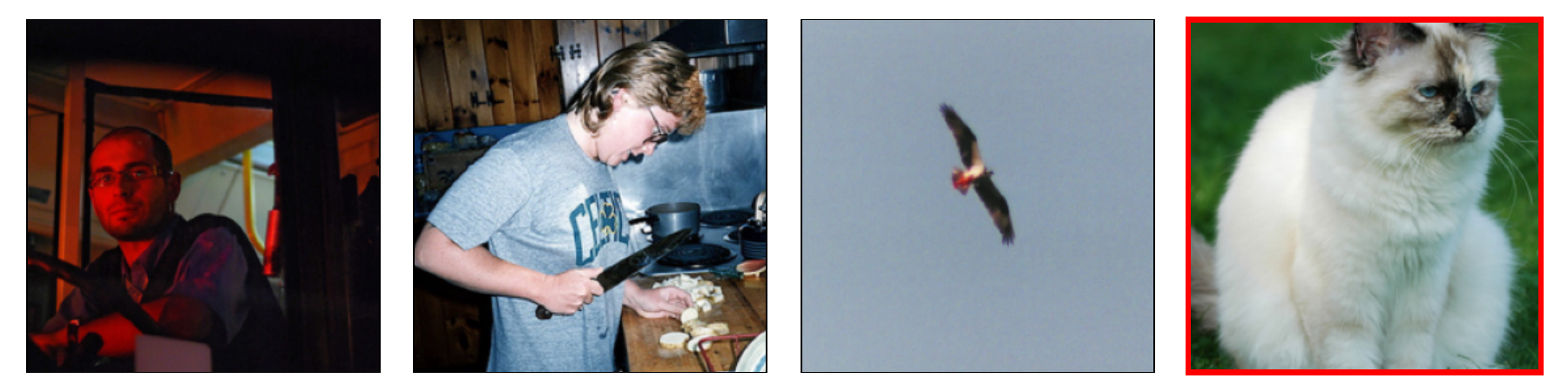}\\[2pt]
C2FA\\
\includegraphics[width=0.85\columnwidth]{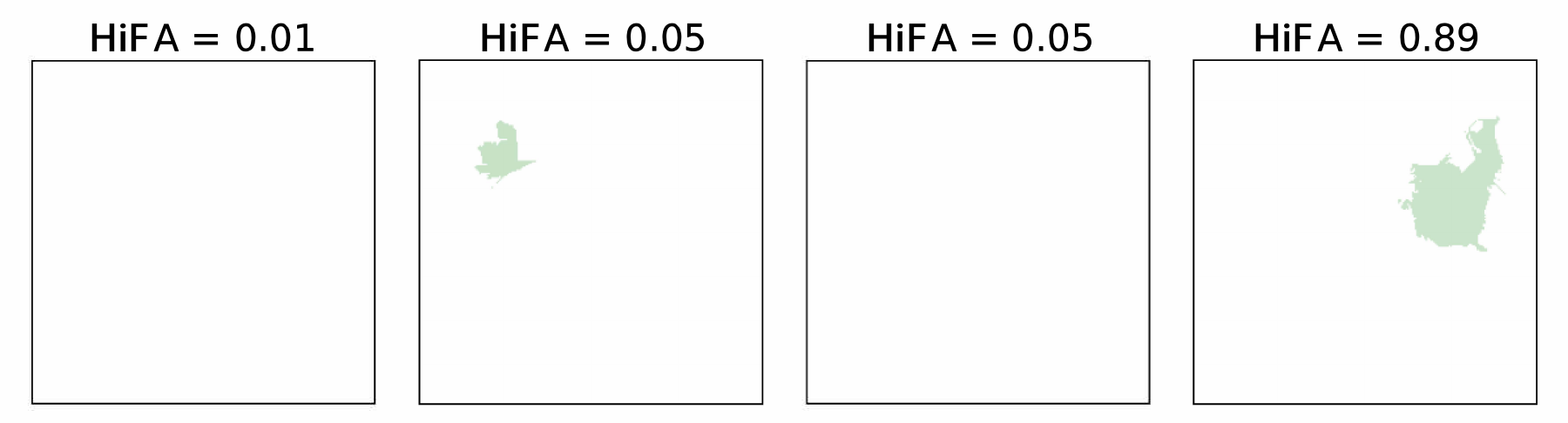} \\[2pt]
LIME \\
\includegraphics[width=0.85\columnwidth]{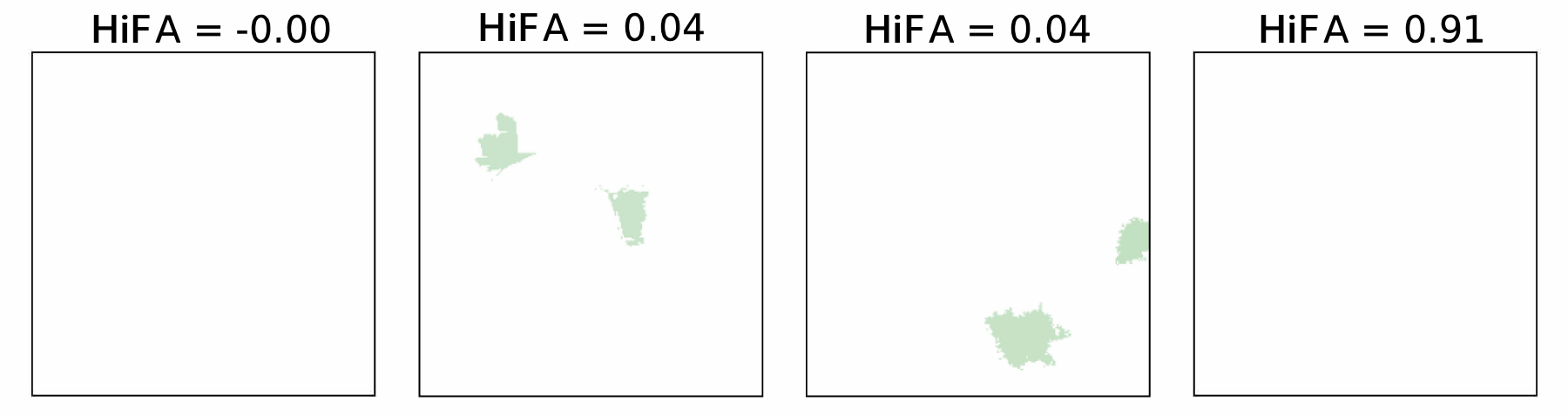}
\caption{Example of the estimated HiFAs and LoFAs on the image classification task when $N_{\rm{H}}=20$ and $N_{\rm{L}}=50$.
The input is shown on the first row, where the image with the red border is the positive instance.
The LoFAs of super-pixels estimated by the proposed method and LIME are shown in the second and third rows, respectively. In each case, super-pixels are highlighted in green, with the intensity of the green color indicating the magnitude of the LoFA.
}
\label{fig:experiment:image:example}
\end{figure}

\paragraph{Results.}
Figure~\ref{fig:experiment:vocbag:HiFA} shows the NDCG and deletion scores of the estimated HiFAs over various numbers of perturbations for the LoFAs, $N_{\rm{L}}$, where we fixed the number of perturbations for the HiFAs, $N_{\rm{H}} = 5$.
We found that the proposed method (C2FA) consistently achieved the best NDCG and deletion scores, and the superiority of the proposed method is especially noticeable when $N_{\rm{L}}$ is small.
Although BU-LIME improved the scores as $N_{\rm{L}}$ increased, the scores were still lower than those of the proposed method.
Since the other comparing methods estimate the HiFAs without the effects of the LoFAs, their scores were constant regardless of the value of $N_{\rm{L}}$.
In Appendix~\ref{sec:appendix:image:IA}, we show that similar results were obtained in terms of the insertion metric.
In addition, when we fixed $N_{\rm{H}} = 20$, the methods other than BU-LIME equally achieved the highest NDCG and insertion scores regardless of $N_{\rm{L}}$ because $N_{\rm{H}}$ was sufficiently large to estimate the HiFAs accurately.

Figure~\ref{fig:experiment:vocbag:LoFA} shows the AUROC and deletion scores of the estimated LoFAs over various values of $N_{\rm{L}}$ where we fixed $N_{\rm{H}} = 20$.
When $N_{\rm{L}}$ is small, we found that the proposed method achieved significantly better AUROC and deletion scores.
In particular, the AUROC score of the proposed method at $N_{\rm{L}} = 50$ was comparable to that of the second-best methods, LIME, MILLI, and BU-LIME, at $N_{\rm{L}} = 150$, and the deletion score of the proposed method at $N_{\rm{L}} = 50$ was much the same as that of the second-best methods at $N_{\rm{L}} = 100$.
These results show that the proposed method is very efficient in terms of the number of queries to the model $f$, thanks to the simultaneous estimation of the HiFAs and LoFAs.

Figure~\ref{fig:experiment:vocbag:consistency} shows the consistency scores and the agreement scores of MIHL over various values of $N_{\rm{L}}$ where we fixed $N_{\rm{H}} = 20$.
Here, the consistency scores of BU-LIME, TD-LIME, and TD-MILLI are always zero by definition.
We found that the consistency scores of LIME and MILLI were worse because they estimated the HiFAs and LoFAs separately.
On the other hand, those of the proposed method were nearly zero, which means that the estimated HiFAs and LoFAs satisfied the consistency property.
With the agreement scores of MIHL, we found that the proposed method outperformed the other methods regardless of the values of $N_{\rm{L}}$, and the differences in the scores were especially noticeable for smaller $N_{\rm{L}}$ values (i.e., $N_{\rm{L}} \leq 150$).

We visualize an example of the estimated HiFAs and LoFAs by the proposed method and the best-performing baseline, LIME, in Figure~\ref{fig:experiment:image:example}.
Here, we only display the LoFAs larger than 0.1 for clarity.
The figure shows that the proposed method assigned a high LoFA to the super-pixel in the high-level feature which was the positive instance and had the highest HiFA (HiFA = 0.89), although LIME assigned high LoFAs to the super-pixels in the negative instances.
The critical difference between the two methods is whether the HiFAs and LoFAs are estimated simultaneously or separately.
Since both the proposed method and LIME assigned the highest HiFA to the positive instance correctly, the result indicates that estimating the HiFAs and LoFAs simultaneously is effective.
Similar results were obtained in other examples shown in Appendix~\ref{sec:appendix:image:example}. %

\subsection{Text Classification Using Language Models}\label{sec:experiment:text}

Another practical application of the proposed method is to explain the attributions of sentences and the words in text classification with language models.

\paragraph{Dataset.}
We constructed a dataset in which the validation and test sets contain 500 and 1,000 product reviews, respectively, randomly sampled from the Amazon reviews dataset~\cite{Zhang2015-sm}, respectively.
Each sample consists of multiple sentences (high-level features), where each sentence is a sequence of words (low-level features).
The label of each sample corresponds to the review's sentiment polarity (positive or negative).

\paragraph{Black-box Model.}
We used BERT~\cite{DBLP:journals/corr/abs-1810-04805} fine-tuned on the original Amazon reviews dataset, which is provided on Hugging Face~\cite{Harel-Canada_undated-qi}.
The test accuracy of the model is 0.947.
When masking a word in the input to generate perturbed inputs, we replaced the word with the predefined mask token \texttt{[MASK]}.
Similarly, when masking a sentence, we replaced all the words in the sentence with the mask token.

\paragraph{Quantitative Evaluation.}
Because no ground-truth labels for HiFAs and LoFAs are available in the dataset, we evaluated the estimated HiFAs and LoFAs only in terms of faithfulness and consistency, as in Section~\ref{sec:experiment:image}.

\begin{figure}[t]
\centering
\begin{subfigure}{0.23\textwidth}
  \includegraphics[width=\textwidth]{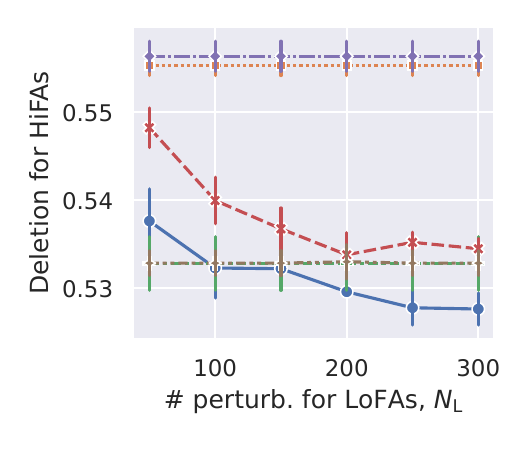}\\
  \includegraphics[width=\textwidth]{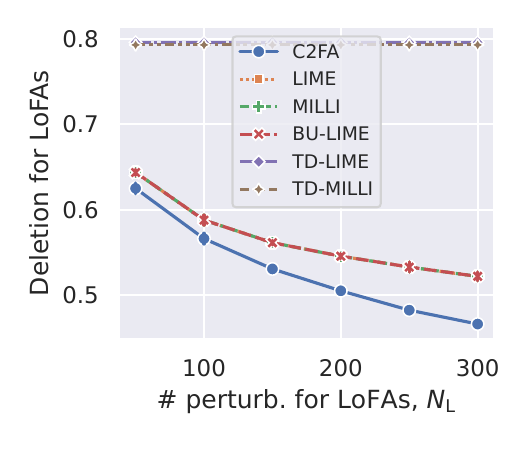}
  \caption{}
  \label{fig:experiment:text:faithfulness}
\end{subfigure}
\begin{subfigure}{0.23\textwidth}
  \includegraphics[width=\textwidth]{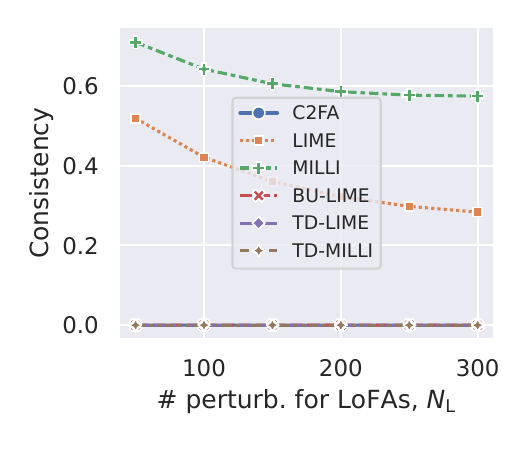}\\
  \includegraphics[width=\textwidth]{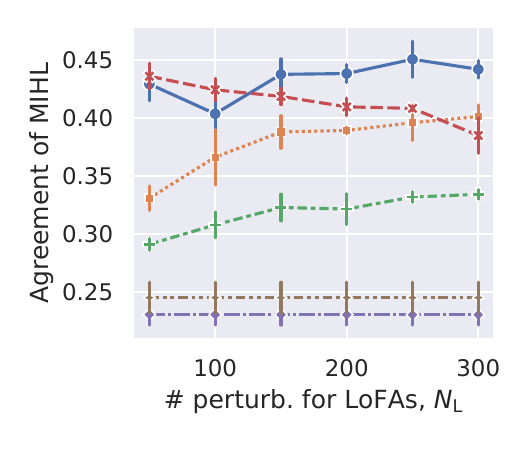}
  \caption{}
  \label{fig:experiment:text:consistency}
\end{subfigure}
\caption{
  Quantitative evaluation on the text classification task.
  (a) Deletion scores of the HiFAs and LoFAs (lower is better).
  (b) Consistency scores (lower is better) and agreement scores of MIHL (higher is better).
}
\label{fig:experiment:text}
\end{figure}

\begin{figure}[t]
\scriptsize
\centering
{\bf Input (bag of sentences)}\\
\begin{tabular}{rp{20em}}
\hline
S1: & do \colorbox{pink}{not} \colorbox{pink}{buy} \colorbox{pink}{this} product .\\
S2: & they break too easily \colorbox{pink}{and} when you want to \colorbox{pink}{replace} them \colorbox{pink}{it} \colorbox{pink}{is} labeled \colorbox{pink}{poorly} . \\
\hline
\end{tabular}
\\[2mm]
C2FA\\
\includegraphics[width=0.7\columnwidth]{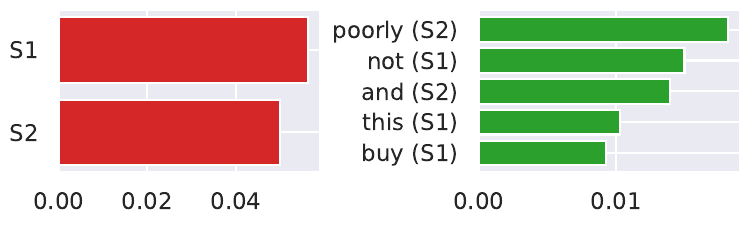}\\
BU-LIME\\
\includegraphics[width=0.7\columnwidth]{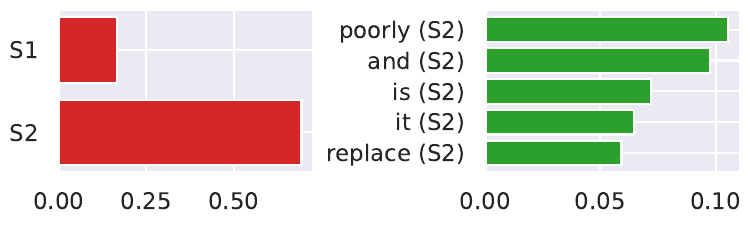} 
\caption{Example of the estimated HiFAs and LoFAs for a negative review text when $N_{\rm{H}}=50$ and $N_{\rm{L}}=50$.
The review text is shown at the top, and the HiFAs (left) and the **top-5 LoFAs (right) estimated by each method are shown at the bottom. 
Here, the words highlighted with a pink background in the review correspond to those top-5 LoFAs in the chart
}
\label{fig:experiment:text:example}
\end{figure}

\paragraph{Results.}
Figure~\ref{fig:experiment:text:faithfulness} shows the deletion scores of the estimated HiFAs and LoFAs over various values of $N_{\rm{L}}$ where we fixed $N_{\rm{H}} = 5$ and $50$, respectively.
With the deletion scores of the HiFAs, although the scores of the proposed method were equal to or worse than those of MILLI and TD-MILLI at $N_{\rm{L}} \leq 150$, the proposed method achieved the best deletion score at $N_{\rm{L}} \geq 200$.
We found that in this task, the LIME-based methods, including the proposed method, were worse than the MILLI-based methods at small $N_{\rm{L}}$ values.
As $N_{\rm{L}}$ increased, the proposed method benefited from the consistency constraints and became the only LIME-based method that outperformed the MILLI-based methods.
In Appendix~\ref{sec:appendix:text:quantitative}, we show that similar results were obtained in terms of the insertion metric, and when we fixed $N_{\rm{H}} = 50$, the scores did not change regardless of the values of $N_{\rm{L}}$ because $N_{\rm{H}}$ was sufficiently large to estimate the HiFAs accurately. %
With the deletion scores of the LoFAs, the proposed method outperformed the other methods regardless of the values of $N_{\rm{L}}$.

Figure~\ref{fig:experiment:text:consistency} shows the consistency scores and the agreement scores of MIHL over various values of $N_{\rm{L}}$ where we fixed $N_{\rm{H}} = 50$.
Again, in this task, the consistency scores of the proposed method were nearly zero regardless of the values of $N_{\rm{L}}$.
With the agreement scores of MIHL, the proposed method maintained high scores regardless of the values of $N_{\rm{L}}$, although the scores of BU-LIME were slightly better than the proposed method at $N_{\rm{L}} \leq 100$.

Figure~\ref{fig:experiment:text:example} shows an example of the HiFAs and LoFAs estimated by the proposed method and the second-best method, BU-LIME. 
In the example, we fixed $N_{\rm{H}} = 50$ and $N_{\rm{L}} = 50$; that is, $N_{\rm{L}}$ is insufficient to estimate the LoFAs accurately.
We found that although the baseline method assigned higher LoFAs to the words in the second sentence (S2), the proposed method assigned higher LoFAs to the words in the first sentence (S1).
This is because the proposed method regularizes the LoFAs via the consistency constraints, exploiting the fact that S1 has a high HiFA.
Other examples are shown in Appendix~\ref{sec:appendix:text:example}.  %

\section{Conclusion}\label{sec:conclusion}
We hypothesized that the consistency property, which is naturally derived from the characteristics of surrogate models, is essential for producing explanations that are accurate, faithful, and consistent between HiFAs and LoFAs. 
This property also enables the method to require fewer queries to the model.
To address this, we proposed a model-agnostic local explanation method for nested-structured inputs, that is capable of explaining two-level feature attributions.
Our experiments on image and text classification tasks demonstrated that the proposed method can efficiently generate high-quality explanations while using fewer queries.
Future work includes extending the method to handle more complex nested structures and exploring additional application domains.

\appendix

\section{Optimization Algorithm}\label{sec:appendix:optimization}

Our objective to estimate the HiFAs and LoFAs simultaneously is as follows:

\begin{align}
\hat{\balpha},\hatbbetad &= \argmin_{\balpha,\bbetad} \Lin(\balpha) + \Lfe(\bbetad) 
+ \lambda_{\rm{H}} \Omega_{\rm{H}}(\balpha) + \lambda_{\rm{L}}\Omega_{\rm{L}}(\bbetad)\nonumber\\
&\quad\mathrm{s.t.}\quad \alpha_j = \sum_{d=1}^{D_j} \beta_{jd} \quad (\forall j \in [J]).
\label{eq:appendix:objective1}
\end{align}

We solve the optimization based on the alternating direction method of multipliers (ADMM)~\cite{Boyd_undated-bc}. 
By introducing auxiliary variables $\bbalpha \in \mathbb{R}^J$ and $\bbbetad \in \mathbb{R}^{\Dd}$ and Lagrange multipliers $\bv_1 \in \mathbb{R}^{J}$, $\bv_2 \in \mathbb{R}^{\Dd}$, and $\bv_3 \in \mathbb{R}^{J}$ based on the ADMM manner, our objective is rewritten as follows:
\begin{align}
\label{eq:appendix:objective2}
&\hat{\balpha},\hatbbetad = \argmin_{\balpha,\bbetad} \Lin(\balpha) + \Lfe(\bbetad) + \lambda_{\rm{H}} \Omega_{\rm{H}}(\bbalpha) + \lambda_{\rm{L}}\Omega_{\rm{L}}(\bbbetad) \nonumber\\
&\quad + \bv_1^{\top} \bh_1(\balpha,\bbalpha) + \bv_2^{\top} \bh_2(\bbetad,\bbbetad) + \bv_3^{\top} \bh_3(\balpha,\bbetad) \nonumber\\
&\quad + \frac{\mu_1}{2}\left\{ \| \bh_1(\balpha,\bbalpha) \|^2 + \| \bh_2(\bbetad,\bbbetad) \|^2 \right\} \nonumber\\
&\quad + \frac{\mu_2}{2} \| \bh_3(\balpha,\bbetad) \|^2,
\end{align}
where $\bh_1(\balpha,\bbalpha) = \balpha - \bbalpha$, $\bh_2(\bbeta^{\dagger},\bbbeta^{\dagger}) = \bbeta^{\dagger} - \bbbeta^{\dagger}$, $\bh_3(\balpha,\bbeta^{\dagger}) = \balpha - \bM \bbeta^{\dagger}$.
Here, $\bM \in \{0,1\}^{J \times \Dd}$ is a binary matrix to add up the LoFAs associated with the same high-level feature where we set to $M_{jd} = 1$ if the $d$th feature of the concatenated input $\bxd$ belongs to the $j$th high-level feature $\bx_j$, and $M_{jd} = 0$ otherwise.
The hyperparameters $\mu_1 \geq 0$ and $\mu_2 \geq 0$ are the penalty parameters for the regularization and the consistency constraint, respectively.

The optimization of~\bref{eq:appendix:objective2} is performed by alternating the updates of the variables iteratively.
We summarize the optimization algorithm using $\ell_2$ regularization for $\Omega_{\rm{H}}$ and $\Omega_{\rm{L}}$ in Algorithm~\ref{alg:appendix:optimization}.
Here, $\bI_J$ and $\bI_{\Dd}$ are identity matrices of size $J$ and $\Dd$, respectively, and in Line 9, the variables at the zeroth step are initialized with zero.
The algorithm is terminated when $\| \bbalpha^{t-1} - \bbalpha^{t} \|^2 + \| \bbbetad{}^{t-1} - \bbbetad{}^{t} \|^2 < \epsilon_1$ and $\| \bh_1(\balpha^{t},\bbalpha^{t}) \|^2 + \| \bh_2(\bbetad{}^{t},\bbbetad{}^{t}) \|^2 + \| \bh_3(\balpha^{t},\bbetad{}^{t}) \|^2 < \epsilon_2$ where $\epsilon_1,\epsilon_2 \geq 0$ are hyperparameters.
The other hyperparameters of the algorithm are $\lambda_{\rm{H}}$, $\lambda_{\rm{L}}$, $\mu_1$, and $\mu_2$.

\begin{algorithm}[t]
  \caption{Estimating consistent two-level feature attributions (C2FA) with $\ell_2$ regularization}
  \label{alg:appendix:optimization}
\begin{algorithmic}[1]
  \STATE Generate binary random matrices $\bZin$ and $\bZfe$
  \STATE Obtain perturbed inputs $\{\tbX^{\rm{H}}_n\}_{n=1}^{N_{\rm{H}}}$ and $\{\tbX^{\rm{L}}_n\}_{n=1}^{N_{\rm{L}}}$ using $\phiin$ and $\phife$
  \STATE Obtain predictions $\tbyin$ and $\tbyfe$ from the perturbed inputs
  \STATE Obtain weight matrices $\bWin$ and $\bWfe$
  \STATE $\bA = ({\bZin}^{\top} \bWin \bZin + (\mu_1 + \mu_2)\bI_J)^{-1}$
  \STATE $\bB = \bA {\bZin}^{\top}\bWin \tbyin$
  \STATE $\bC = ({\bZfe}^{\top} \bWfe \bZfe + \mu_1\bI_{\Dd} + \mu_2\bM^{\top}\bM)^{-1}$
  \STATE $\bD = \bC{\bZfe}^{\top}\bWfe \tbyfe$
  \STATE Initialize $\balpha^{0}$, $\bbalpha^{0}$, $\bbetad{}^{0}$, $\bbbetad{}^{0}$, $\bv_1^{0}$, $\bv_2^{0}$, $\bv_3^{0}$ with zero
  \STATE $t = 0$
  \REPEAT
  \STATE $\balpha^{t+1} = \bB + \bA(\mu_2\bM\bbetad{}^{t} - \mu_1\bbalpha^{t} - \bv_1^{t} - \bv_3^{t})$
  \STATE $\bbalpha^{t+1} = (\mu_1 + 2 \lambda_{\rm{H}})^{-1} (\bv^t_1 + \mu_1 \balpha^{t+1})$
  \STATE $\bbetad{}^{t+1} = \bD + \bC(\bM^{\top}\bv_3^{t} + \mu_1\bbbetad{}^{t} + \mu_2\bM^{\top}\balpha^{t+1} - \bv_2^{t})$
  \STATE $\bbbetad{}^{t+1} = (\mu_1 + 2 \lambda_{\rm{L}})^{-1} (\bv^t_2 + \mu_1\bbetad{}^{t+1})$
  \STATE $\bv_1^{t+1} = \bv_1^{t} + \mu_1(\balpha^{t+1} - \bbalpha^{t+1})$
  \STATE $\bv_2^{t+1} = \bv_2^{t} + \mu_1(\bbetad{}^{t+1} - \bbbetad{}^{t+1})$
  \STATE $\bv_3^{t+1} = \bv_3^{t} + \mu_2(\balpha^{t+1} -\bM\bbetad{}^{t+1})$
  \STATE $t = t + 1$
  \UNTIL{stop criterion is met}
  \STATE {\bfseries return:} $\bbalpha^{t}$, $\bbbetad{}^{t}$
\end{algorithmic}
\end{algorithm}

\section{Computational Time Complexity}\label{sec:appendix:complexity}

The computational time complexity of Algorithm~\ref{alg:appendix:optimization} is split into three parts.
The first part is the predictions for the perturbed inputs (Line 3), which is $O((N_{\rm{H}}+N_{\rm{L}})Q)$ where $Q$ is the computational cost of the model $f$ in prediction.
The second part is the pre-computation before the iterations (Lines 4--8), which is $O(J^3 + \Dd{}^3 + J^2N_{\rm{H}} + JN_{\rm{H}}^2 + \Dd{}^2 N_{\rm{L}} + \Dd N_{\rm{L}}^2)$.
The third part is the iterations (Lines 11--20), which is $O(TJ^2\Dd + TJ\Dd{}^2)$ where $T$ is the number of iterations.
Compared with estimating the HiFAs and LoFAs by solving~\bref{eq:proposed:HiFA} and~\bref{eq:proposed:LoFA} separately,
the third part is an additional computational cost in the proposed method.
However, because one wants to execute the model on low-resource devices and cloud services, $Q$ is often large; consequently, the first part could be dominant.
Therefore, estimating the HiFAs and LoFAs accurately with small $N_{\rm{H}}$ and $N_{\rm{L}}$, i.e., small amounts of perturbed inputs, is crucial in practical situations.
In the experiments in Section~\ref{sec:experiment}, we demonstrate that the proposed method can estimate high-quality HiFAs and LoFAs even when $N_{\rm{H}}$ and $N_{\rm{L}}$ are small.
In Appendix~\ref{sec:appendix:text:time}, we show that the actual computational time scales linearly with $N_{\rm{H}}$ and $N_{\rm{L}}$ as with estimating the HiFAs and LoFAs separately.

\section{Experiments on Image Classification in Multiple Instance Learning}\label{sec:appendix:image}

\subsection{Detailed Description of Dataset}\label{sec:appendix:image:dataset}
We constructed an MIL dataset from the Pascal VOC semantic segmentation dataset~\cite{Everingham15} that allows us to evaluate the estimated HiFAs and LoFAs with the ground-truth instance- and pixel-level labels.
With the training subset of the dataset, each bag contains **three to five images**, randomly selected from the Pascal VOC dataset.
Here, low-level features correspond to regions (super-pixels) of each image, which are obtained by the quick shift algorithm~\cite{Vedaldi2008-us}.
Each bag is labeled positive if at least one image in the bag is associated with the ``cat'' label; otherwise the bag if negative.
Also, each image pixel is labeled positive if the pixel is associated with ``cat'' label and negative otherwise.
We used the instance- and pixel-level supervision only for evaluation.
Similarly, we constructed validation and test subsets whose samples contain images from the training and test subsets of the Pascal VOC, respectively.
The number of samples in training, validation, and test subsets is 5,000, 1,000, and 2,000, respectively, and the positive and negative samples ratio is equal.

\subsection{Implementation Details}\label{sec:appendix:image:implementation}
We defined the DeepSets permutation-invariant model~\cite{Zaheer2017-in} as a black-box model $f$ to be explained.
According to \cite{Zaheer2017-in}, the model $f$ comprises two components: a representation function that transforms each instance, $\phi$, and a non-linear network that produces predictions from the extracted representation, $\rho$.
We used ResNet-50~\cite{He2016-kb} pre-trained on ImageNet as the representation function $\phi$ and two-layer multi-layer perceptron (MLP) as the non-linear network $\rho$.
Here, in $\rho$, we used the ReLU activation function for the first layer and the softmax function for the second layer.
Also, the number of hidden units in the MLP was set to 1,024.
The model first extracts the representation of each instance using $\phi$, then adds them up into a single representation, and finally, outputs a prediction by applying $\rho$ to the aggregated single representation.
We trained the model using our MIL image classification dataset with Adam optimizer~\cite{Kingma2014-qn} with a learning rate of 0.001, a batch size of 32, and a maximum epoch of 300.
The test accuracy of the model was 0.945.

The hyperparameters of C2FA, $\lambda_{\rm{H}}$, $\lambda_{\rm{L}}$, and $\mu_1$, were tuned using the validation subset of each dataset within the following ranges: $\lambda_{\rm{H}}, \lambda_{\rm{L}} \in \{0.1, 1 \}$, and $\mu_2 \in \{0.001,0.01,0.1\}$.
The remaining hyperparameters were set to $\mu_1 = 0.1$ and $\epsilon_1 = \epsilon_2 = 10^{-4}$.
The experiments were conducted on a server with an Intel Xeon Gold 6148 CPU and an NVIDIA Tesla V100 GPU.

\subsection{Detailed Description of Quantitative Evaluation Metrics}\label{sec:appendix:image:evaluation}
We assessed the estimated HiFAs and LoFAs in terms of correctness, faithfulness, and consistency.
The correctness is evaluated using the ground-truth instance- and pixel-level labels.
Following the evaluation in the MIL study~\cite{Early2022-oc}, we evaluated the estimated HiFAs using normalized discounted cumulative gain (NDCG).
For the estimated LoFAs, as with the evaluation of the LoFAs for single image classification~\cite{Sampaio2023-ji}, we evaluated them as the predictions of the pixel-level labels by the area under ROC curve (AUROC) in the binary semantic segmentation manner.
For faithfulness, we assessed whether the estimated HiFAs and LoFAs are faithful to the behaviors of the model $f$ using insertion and deletion metrics~\cite{Petsiuk2018-uj}.
The insertion and deletion metrics evaluate the change in the predictions of the model $f$ when features deemed important in the LoFAs are gradually added and removed from the sample, respectively.
In our experiments, we gradually add and remove the low-level features across all the high-level features in descending order of their LoFAs.
Also, for the HiFAs, we add and remove the high-level features instead of the low-level ones, respectively.
In terms of the consistency evaluation, we used the following two metrics.
The first one is the consistency between the estimated HiFAs and LoFAs, which is calculated with $\| \balpha - \bM \bbetad \|^2$ used to calculate the penalty for the consistency constraints in~\bref{eq:appendix:objective2}.
The second one is the agreement of the most important high- and low-level feature (MIHL), which is calculated by the ratio that the high-level feature of the highest HiFA is identical to the one associated with the low-level feature of the highest LoFA.  

We evaluated the above metrics using only the samples with the positive bag label because we could not evaluate the correctness of those with the negative bag label.
We ran the evaluations three times with different random seeds and reported the average scores and their standard deviation.

\subsection{Additional Quantitative Evaluation}\label{sec:appendix:image:IA}

\begin{figure}[t]
\centering
\includegraphics[width=0.9\columnwidth]{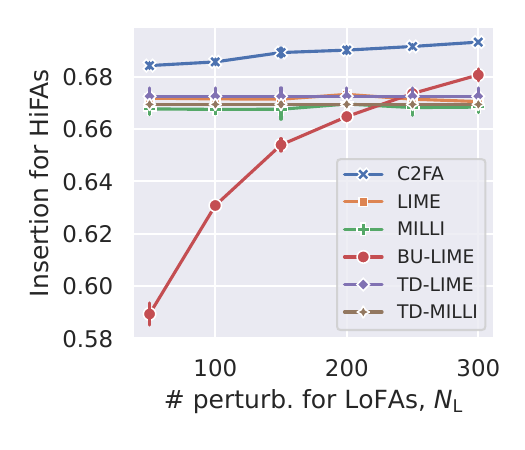}
\vspace{-3mm}
\caption{
  Insertion scores (higher is better) of the estimated HiFAs on the image classification task.
  The error bars represent the standard deviations of the scores over three runs with different random seeds.
}
\label{fig:appendix:image:insertion}
\end{figure}

Figure~\ref{fig:appendix:image:insertion} shows the insertion scores of the estimated HiFAs over various numbers of perturbations for the LoFAs, $N_{\rm{L}}$, where we fixed the number of perturbations for the HiFAs, $N_{\rm{H}} = 5$.
As with the deletion scores in Figure~\ref{fig:experiment:vocbag:HiFA}, the proposed method consistently achieved the best insertion scores.

\begin{figure*}[t]
\centering
\begin{minipage}{0.25\textwidth}
  \centering
  \includegraphics[width=\columnwidth]{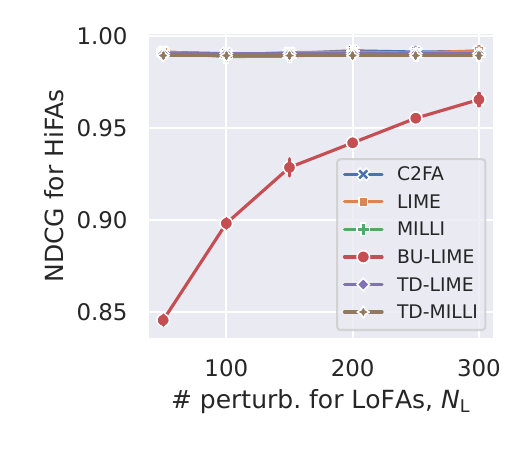}
\end{minipage}
\begin{minipage}{0.25\textwidth}
  \centering
  \includegraphics[width=\columnwidth]{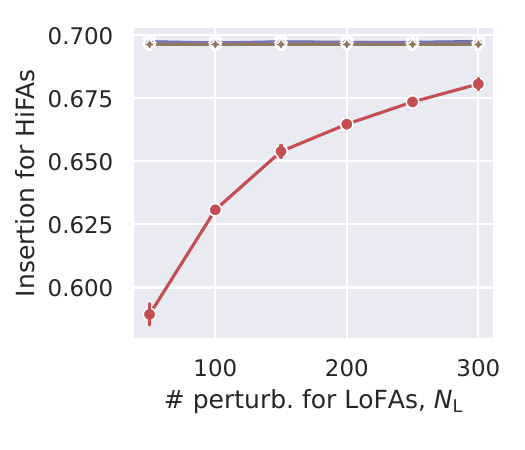}
\end{minipage}
\begin{minipage}{0.25\textwidth}
  \centering
  \includegraphics[width=\columnwidth]{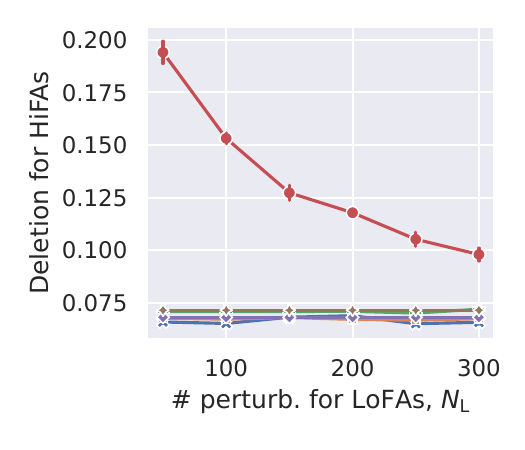}
\end{minipage}
\caption{
  NDCG, insertion, and deletion scores of the estimated HiFAs on the image classification task when the number of perturbed instances $N_{\rm{H}}$ is 20.
}
\label{fig:appendix:image:IA20}
\end{figure*}

Figure~\ref{fig:appendix:image:IA20} shows the NDCG, insertion, and deletion scores of the estimated HiFAs over various $N_{\rm{L}}$, where we fixed $N_{\rm{H}} = 20$.
This result shows that the methods other than BU-LIME equally achieved the highest NDCG and insertion scores regardless of $N_{\rm{L}}$ because $N_{\rm{H}}$ was sufficiently large to estimate the HiFAs accurately.

\subsection{Additional Examples of Estimated Feature Attributions}\label{sec:appendix:image:example}

\begin{figure*}[t]
\begin{minipage}{0.46\textwidth}
\scriptsize
\centering
Input (bag of images)\\
\includegraphics[width=\columnwidth]{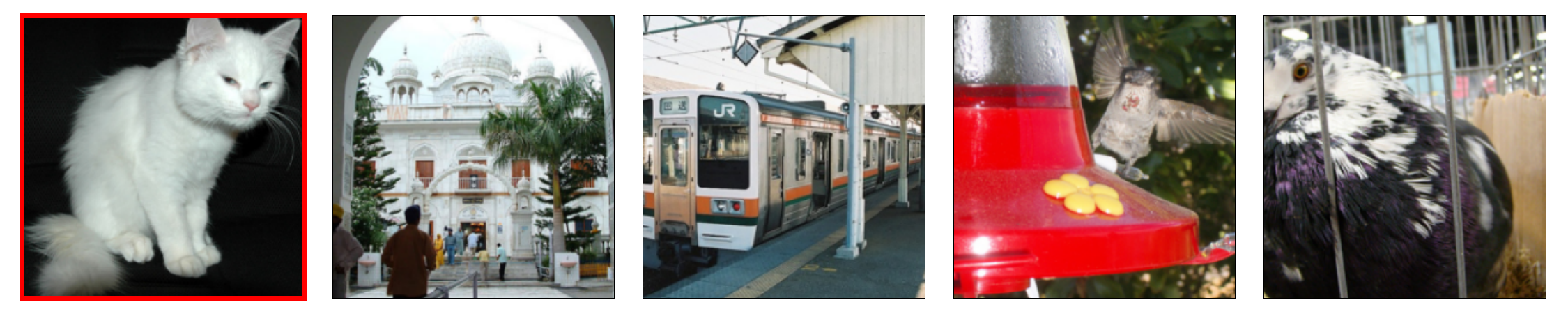}\\[2pt]
C2FA\\
\includegraphics[width=\columnwidth]{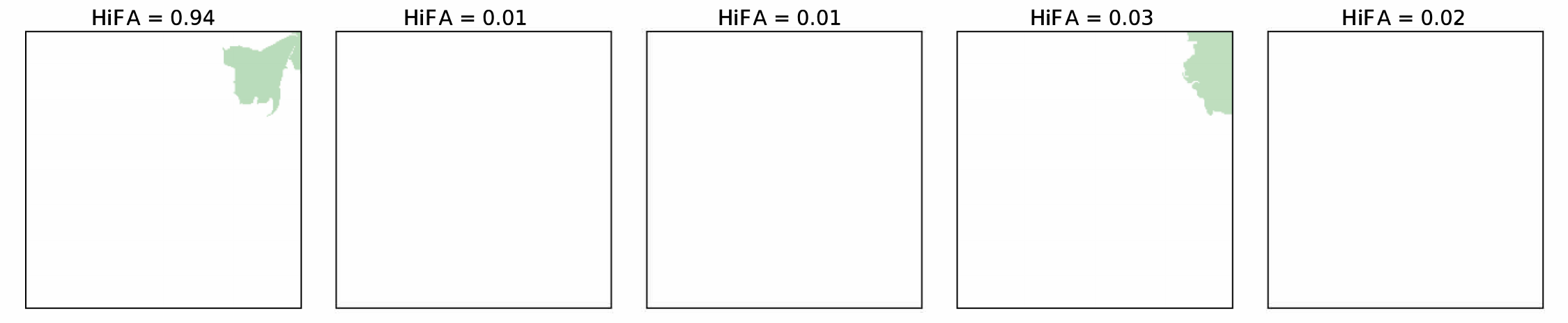}\\[2pt]
LIME\\
\includegraphics[width=\columnwidth]{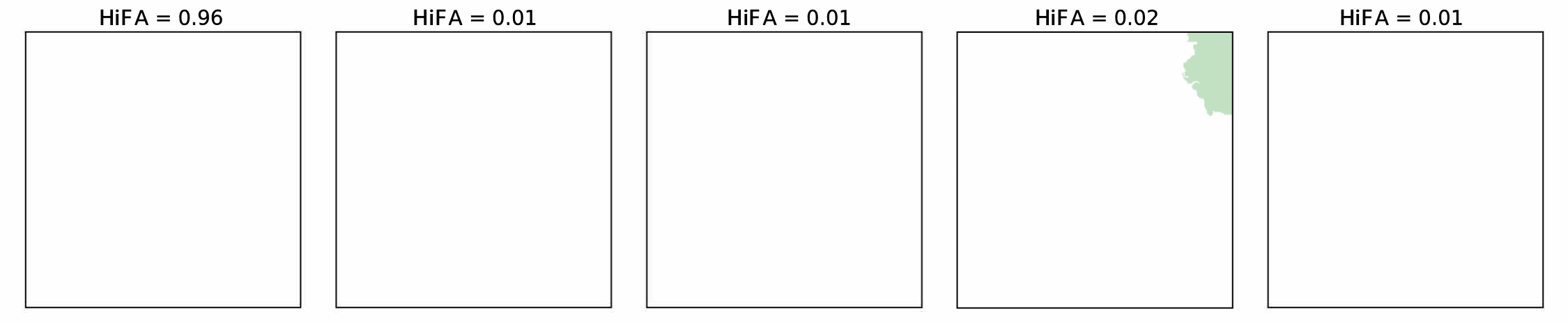}
\end{minipage}
\hfill
\begin{minipage}{0.46\textwidth}
\scriptsize
\centering
Input (bag of images)\\
\includegraphics[width=\columnwidth]{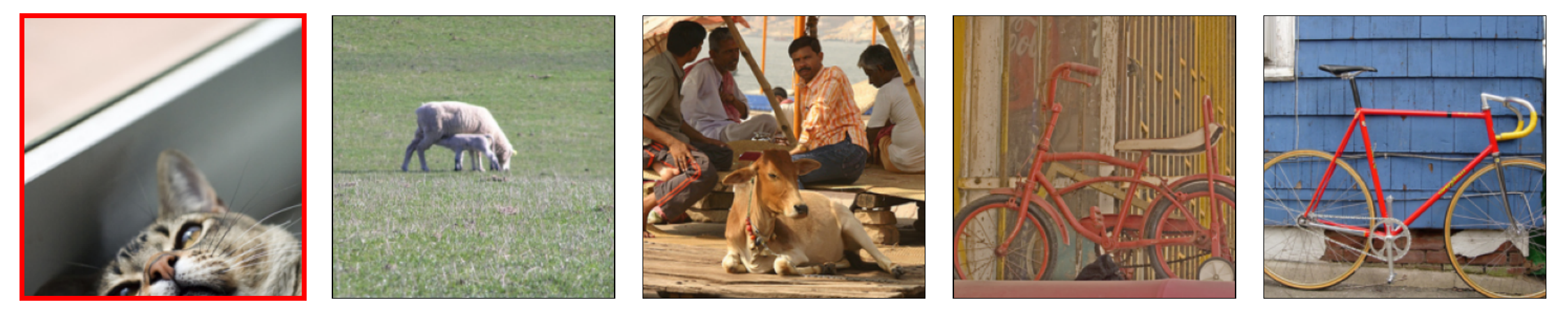}\\[2pt]
C2FA\\
\includegraphics[width=\columnwidth]{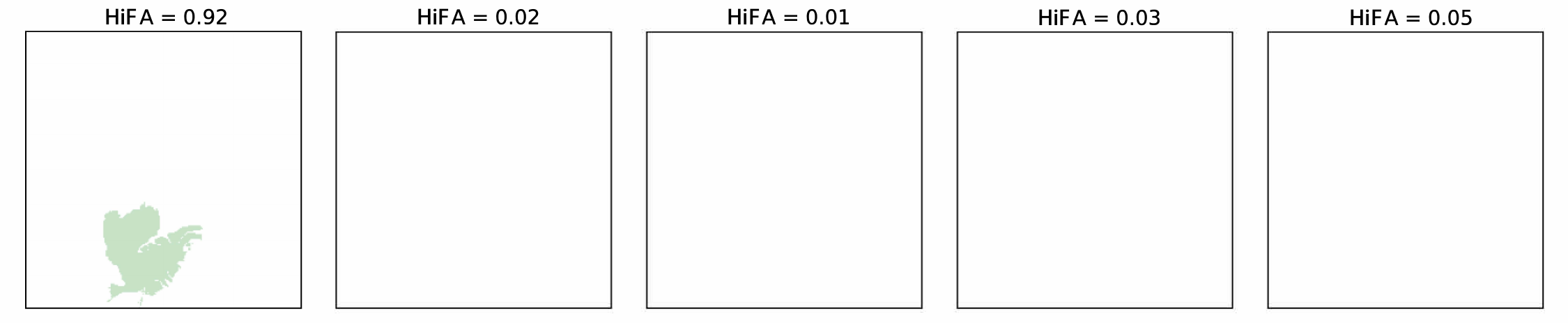} \\[2pt]
LIME\\
\includegraphics[width=\columnwidth]{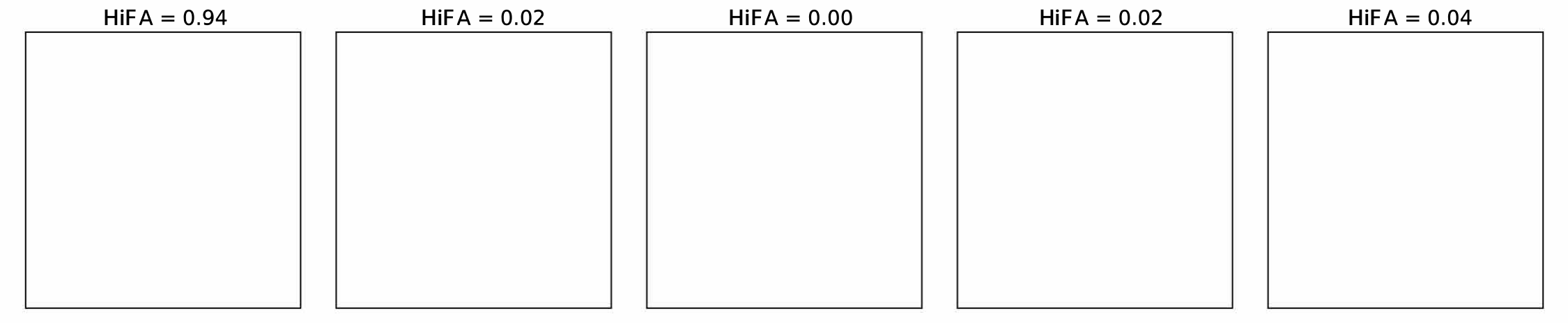}
\end{minipage}
\caption{Two additional examples of the estimated HiFAs and LoFAs in the image classification task when $N_{\rm{H}}=20$ and $N_{\rm{L}}=50$.
The input is shown on the first row, where the image with the red border is the positive instance.
The HiFAs of super-pixels estimated by the proposed method (C2FA) and LIME are shown on the second and third rows, respectively, where the intensity of the green color indicates the magnitude of the LoFA.
Also, the score at the top of each subplot indicates the value of the estimated HiFA for the instance. 
}
\label{fig:appendix:image:example}
\end{figure*}

Figure~\ref{fig:appendix:image:example} shows additional examples of the estimated HiFAs and LoFAs by the proposed method and the best-performing baseline, LIME, on the image classification tasks.
In the setting where $N_{\rm{H}}=20$ and $N_{\rm{L}}=50$, the HiFAs tend to be estimated accurately, and the LoFAs tend to be estimated inaccurately because $N_{\rm{L}}=50$ is small.
Therefore, LIME tended to assign high LoFAs to incorrect regions.
On the other hand, the proposed method was able to assign high LoFAs to correct regions by complementing the insufficiency of $N_{\rm{L}}$ with the accurate HiFAs.

\section{Experiments on Text Classification Using Language Models}\label{sec:appendix:text}

\subsection{Additional Quantitative Evaluation}\label{sec:appendix:text:quantitative}

\begin{figure}[t]
\centering
\begin{minipage}{0.45\columnwidth}
  \centering
  \includegraphics[width=\columnwidth]{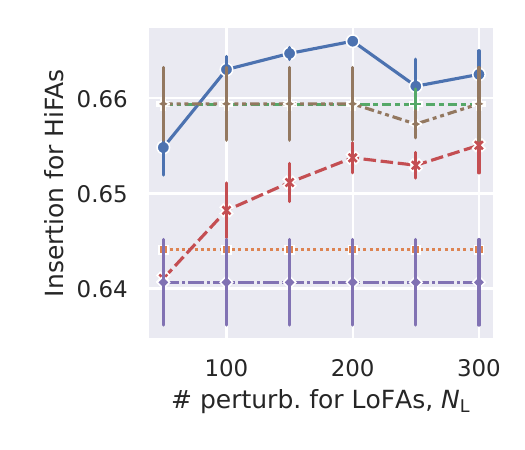}
\end{minipage}
\hspace{2mm}
\begin{minipage}{0.45\columnwidth}
  \centering
  \includegraphics[width=\columnwidth]{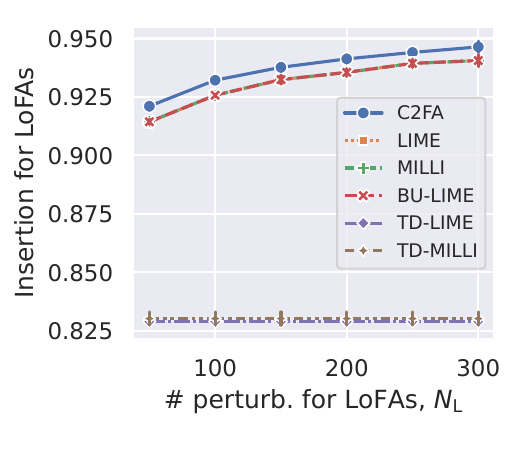}
\end{minipage}
\caption{
  Insertion scores of (left) the estimated HiFAs and (right) the estimated LoFAs on the text classification task (higher is better).
}
\label{fig:appendix:text:insertion}
\end{figure}

\begin{figure}[t]
\centering
\begin{minipage}{0.45\columnwidth}
  \centering
  \includegraphics[width=\columnwidth]{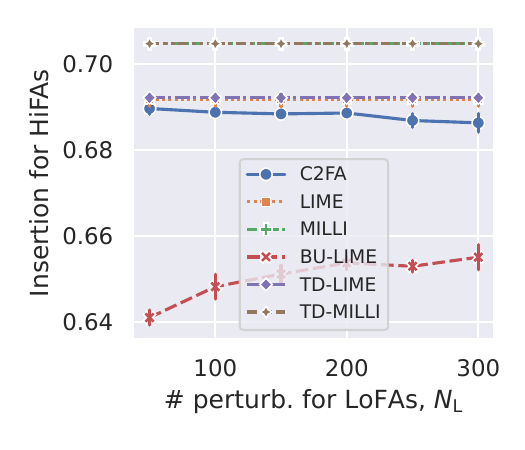}
\end{minipage}
\hspace{2mm}
\begin{minipage}{0.45\columnwidth}
  \centering
  \includegraphics[width=\columnwidth]{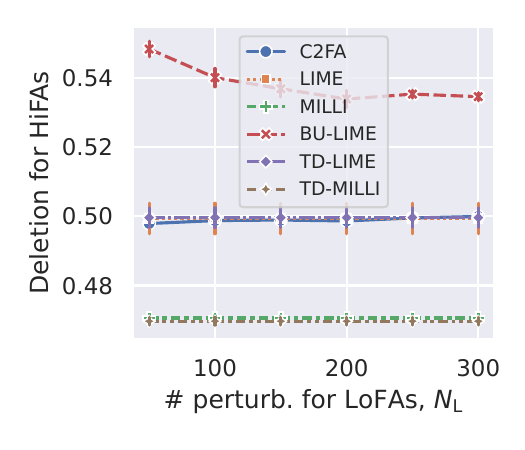}
\end{minipage}
\caption{
  Insertion (higher is better) and deletion (lower is better) scores of the estimated HiFAs on the text classification task when $N_{\rm{H}} = 50$.
}
\label{fig:appendix:text:IA50}
\end{figure}

Figure~\ref{fig:appendix:text:insertion} shows the insertion scores of the estimated HiFAs and LoFAs over various values of $N_{\rm{L}}$ where we fixed $N_{\rm{H}} = 5$ and $50$, respectively.
With the insertion scores of the HiFAs, the proposed method became better than the MILLI-based methods as $N_{\rm{L}}$ increased, as with the deletion scores in Figure~\ref{fig:experiment:text:faithfulness}.
With the insertion scores of the LoFAs, the proposed method outperformed the other methods regardless of the values of $N_{\rm{L}}$.

Figure~\ref{fig:appendix:text:IA50} shows the insertion and deletion scores of the estimated HiFAs over various values of $N_{\rm{L}}$ where we fixed $N_{\rm{H}} = 50$.
In the setting where $N_{\rm{H}}$ is sufficiently large, the MILLI-based methods are superior to the LIME-based methods, including the proposed method, on the text classification task.
This result suggests that the better approach for this task would be to formulate the estimators of the HiFAs~\bref{eq:proposed:HiFA} and LoFAs~\bref{eq:proposed:LoFA} with the MILLI-based sample weight kernel and optimize them simultaneously with the proposed consistency constraints.
However, since no study has applied the idea of MILLI for estimating the LoFAs, we left the attempt for future work.

\subsection{Computational Time}\label{sec:appendix:text:time}

\begin{figure}[t]
\centering
\includegraphics[width=0.9\columnwidth]{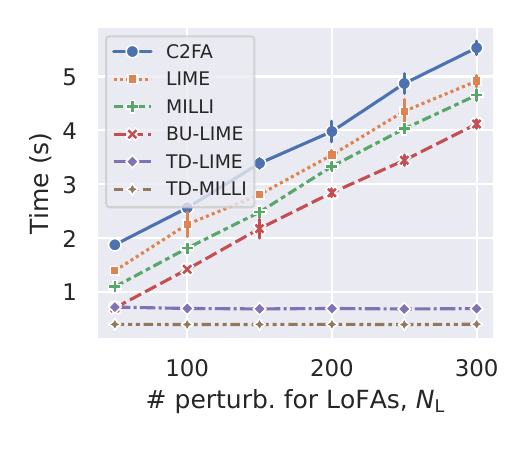}
\vspace{-2mm}
\caption{Average computational time of estimating LoFAs using the proposed method on the text classification task when $N_{\rm{H}} = 50$.}
\label{fig:appendix:text:time}
\end{figure}

Figure~\ref{fig:appendix:text:time} shows the average computational time of estimating the LoFAs using the proposed method on the text classification task when $N_{\rm{H}} = 50$.
From the figure, we can see that the computational time of the proposed method scales linearly with $N_{\rm{L}}$ as with estimating the HiFAs and LoFAs separately.
Here, the computational time of TD-LIME and TD-MILLI is constant against $N_{\rm{L}}$ because they estimate the HiFAs only.

\subsection{Additional Examples of Estimated Feature Attributions}\label{sec:appendix:text:example}

\begin{figure*}[t]
\begin{minipage}[b]{0.48\textwidth}
  \scriptsize
  {\bf Input (bag of sentences)}
  \centering
  \begin{tabular}{rp{22em}}
  \hline
  S1: & \colorbox{pink}{good} quality , especially for \colorbox{pink}{a} long length .\\
  S2: & of \colorbox{pink}{course} component \colorbox{pink}{video} is much better than rca \colorbox{pink}{.} \\
  S3: & looks great on \colorbox{pink}{our} hd \colorbox{pink}{\#\#tv} \colorbox{pink}{.} \\
  \hline
  \end{tabular}
  \\[10pt]
  C2FA\\
  \includegraphics[width=\columnwidth]{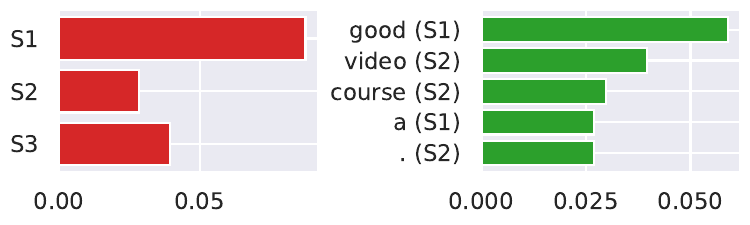}\\[2pt]
  LIME\\
  \includegraphics[width=\columnwidth]{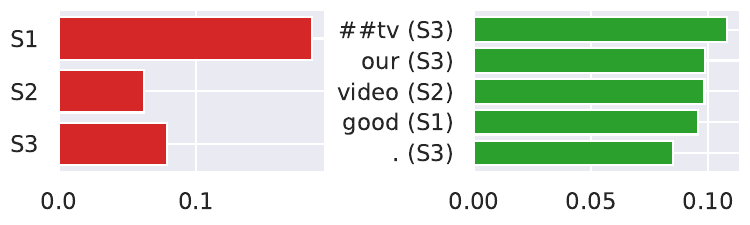} \\[2pt]
  MILLI\\
  \includegraphics[width=\columnwidth]{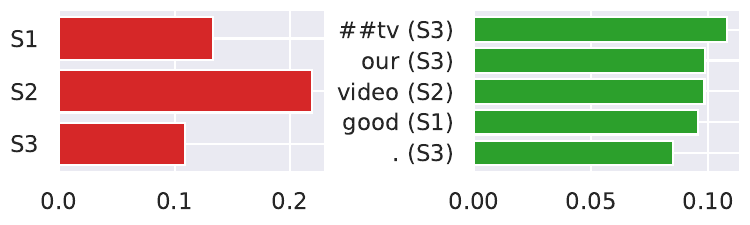} \\[2pt]
  BU-LIME\\
  \includegraphics[width=\columnwidth]{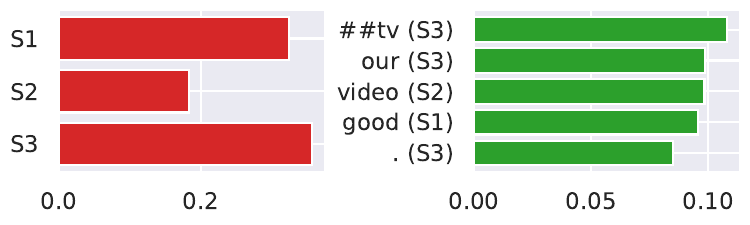}  
\end{minipage}
\hfill
\begin{minipage}[b]{0.48\textwidth}
  \scriptsize
  {\bf Input (bag of sentences)}
  \centering
  \begin{tabular}{rp{22em}}
  \hline
  S1: & 2003 by diana guerrero ( alliance \#\#of \#\#writer \#\#s . com ) detective caroline mab \#\#ry meets lots of lunatic \#\#s on her night shift , but this one with the eye patch is a gem \colorbox{pink}{.}\\
  S2: & he wants to confess , but to what ? \\
  S3: & when he \colorbox{pink}{says} homicide , the \colorbox{pink}{journey} begins . \\
  S4: & the reader travels back in time through his long written confession infused with brief glimpse \#\#s back into the present and the thoughts of our heroine .\\
  S5: & an interesting read , i found \colorbox{pink}{the} description of boy \#\#hood , teen trials , and related events to be vivid and entertaining . \\
  S6: & land of the blind is not your run of the mill detective story . \\
  S7: & \colorbox{pink}{i} \colorbox{pink}{recommend} \colorbox{pink}{it} \colorbox{pink}{.} \\
  \hline
  \end{tabular}
  \\[10pt]
  C2FA\\
  \includegraphics[width=\columnwidth]{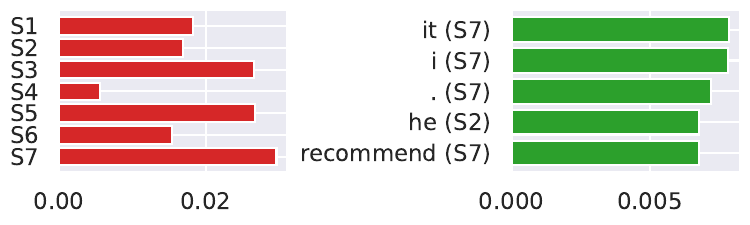}\\[2pt]
  LIME\\
  \includegraphics[width=\columnwidth]{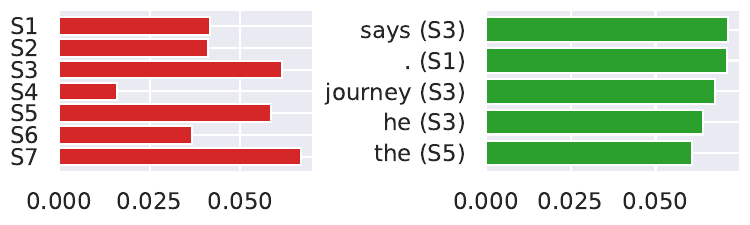} \\[2pt]
  MILLI\\
  \includegraphics[width=\columnwidth]{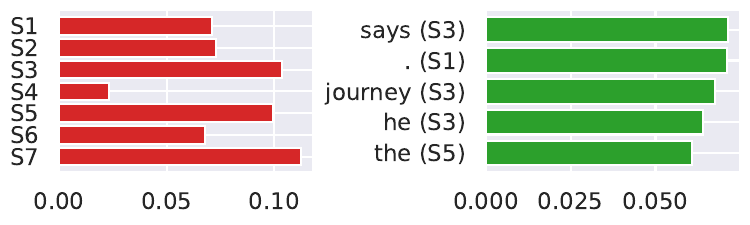} \\[2pt]
  BU-LIME\\
  \includegraphics[width=\columnwidth]{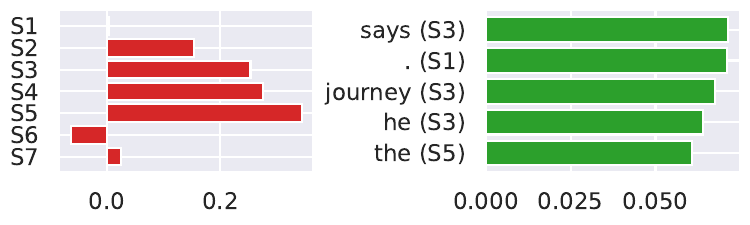}    
\end{minipage}
\caption{Examples of the estimated HiFAs and LoFAs for positive review texts when $N_{\rm{H}}=50$ and $N_{\rm{L}}=50$.
The review text is shown at the top, and the HiFAs (left) and the top-5 highest LoFAs (right) estimated by each method are shown at the bottom.
Here, the words on the pink background in the review text are those appearing in the chart of the LoFAs.
}
\label{fig:appendix:text:example}
\end{figure*}

Figure~\ref{fig:appendix:text:example} shows additional examples of the HiFAs and LoFAs estimated by the proposed method and the three comparing methods on the text classification task.
The result shows that the sentence with the highest HiFA and the sentence associated with the word with the highest LoFA were consistent in the proposed method.
However, the remaining methods did not show such consistency.

\section{Proofs}
\label{sec:appendix:proofs}
\begin{proof}[Proof for Lemma~\ref{lem:existence_of_a_global_minimizer}]
    By assumptions, the loss functions $\Lin(\cdot)$, $\Lfe(\cdot)$, as well as the regularizers $\Omegain(\cdot)$, $\Omegafe(\cdot)$ are convex in their respective parameters.
    Since the objective function $\mathcal{J}$ is convex, lower bounded, and the feasible set $\Gamma^\circ$ is non-empty, closed and convex, by the Weierstrass extreme value theorem, there exists at least one global minimizer $(\balpha^*, \bbeta^*) \in \Gamma^\circ$.
\end{proof}

\begin{proof}[Proof for Lemma~\ref{lem:strong_convexity_and_uniqueness}]
    Suppose, for contradiction, that there are two distinct points $(\balpha^{(1)}, \bbeta^{(1)})$ and $(\balpha^{(2)}, \bbeta^{(2)})$ in $\Gamma^\circ$ such that
    \begin{align*}
        \mathcal{J}(\balpha^{(1)}, \bbeta^{(1)}) &= \min_{(\balpha, \bbeta) \in \Gamma^\circ} \mathcal{J}(\balpha, \bbeta) = \mathcal{J}(\balpha^{(2)}, \bbeta^{(2)}), \\
        (\balpha^{(1)}, \bbeta^{(1)}) &\neq (\balpha^{(2)}, \bbeta^{(2)}).
    \end{align*}
    Consider the midpoint
    \begin{align*}
        \left(\balpha^{(1/2)}, \bbeta^{(1/2)}\right) = \frac{1}{2}(\balpha^{(1)}, \bbeta^{(1)}) + \frac{1}{2}(\balpha^{(2)}, \bbeta^{(2)}).
    \end{align*}
    Since the feasibility constraints $\alpha_j = \sum_d \beta_{j,d}$ are linear in $\balpha$ and $\bbeta$, if each $(\balpha^{(k)}, \bbeta^{(k)})$ satisfies the constraints, then so does their midpoint.
    Hence, $(\balpha^{(1/2)}, \bbeta^{(1/2)}) \in \Gamma^\circ$.
    By strict convexity of the objective $\mathcal{J}(\balpha, \bbeta)$ in its joint argument $(\balpha, \bbeta)$, we have
    \begin{align*}
        \mathcal{J}\left(\balpha^{(1/2)}, \bbeta^{(1/2)}\right) &= \mathcal{J}\left(\frac{1}{2}(\balpha^{(1)}, \bbeta^{(1)}) + \frac{1}{2}(\balpha^{(2)}, \bbeta^{(2)})\right) \\
        &< \frac{1}{2}\mathcal{J}(\balpha^{(1)}, \bbeta^{(1)}) + \frac{1}{2}\mathcal{J}(\balpha^{(2)}, \bbeta^{(2)}) \\
        &= \frac{1}{2}\left\{\mathcal{J}(\balpha^{(1)}, \bbeta^{(1)}) + \mathcal{J}(\balpha^{(2)}, \bbeta^{(2)}) \right\}.
    \end{align*}
    But because each $(\balpha^{(k)}, \bbeta^{(k)})$ is assumed to be a global minimizer, we also have
    \begin{align*}
        \mathcal{J}(\balpha^{(1)}, \bbeta^{(1)}) = \mathcal{J}(\balpha^{(2)}, \bbeta^{(2)}) = \mathcal{J}^*,
    \end{align*}
    where $J^*$ is the common global minimum value.
    Hence,
    \begin{align*}
        \mathcal{J}\left(\balpha^{(1/2)}, \bbeta^{(1/2)}\right) < J^*.
    \end{align*}
    Thus, $\mathcal{J}(\balpha^{(1/2)}, \bbeta^{(1/2)})$ is strictly less than $J^*$, contradicting the global minimality of $J^*$.
\end{proof}

\begin{proof}[Proof for Theorem~\ref{thm:strictly_reduce_objective}]
    Because $\tilde{\balpha}, \tilde{\bbeta}$ violate at least one consistency condition, we can attempt to correct that violation in the space $\Gamma^\circ$.
    Fix $\bbeta$, then $\alpha_j$ must be $\alpha_j = \sum_d \beta_{j,d}$ for feasibility.
    This uniquely determines $\balpha$ is $\bbeta$ is given.
    Hence any $\bbeta \in \Gamma^\circ$ automatically determines $\balpha$.
    Let us define a function
    \begin{align*}
        \Phi(\bbeta) \coloneqq \left(\balpha^\circ(\bbeta), \bbeta \right),
    \end{align*}
    where $\balpha^\circ(\bbeta)$ is given by $\balpha^\circ_j = \sum_d \beta_{j,d}$.
    Then, $\Phi(\bbeta) \in \Gamma^\circ$ for every $\bbeta$.
    By strict convexity assumptions on the objective, one can show that $\mathcal{J}(\Phi(\bbeta))$ as a function of $\bbeta$ is strictly convex.
    The joint minimization problem $\min_{\bbeta}\mathcal{J}(\Phi(\bbeta))$ has a unique solution $\bbeta^*$, and the corresponding $\balpha^* = \balpha^\circ(\bbeta^*)$.
    Thus, $(\balpha^*, \bbeta^*)$ is the unique global minimizer.
    
    Let us form
    \begin{align*}
        (\balpha^t, \bbeta^t) = t(\balpha^*, \bbeta^*) + (1 - t) (\tilde{\balpha}, \tilde{\bbeta}),
    \end{align*}
    for $t \in (0, 1)$.
    Because the constraints are linear, $\Gamma^\circ$ is an affine set, and $(\tilde{\balpha}, \tilde{\bbeta}) \notin \Gamma^\circ$.
    In general, line segment from a feasible point to an infeasible point must leave the feasible region.
    Indeed, it contains strictly interior points that are also infeasible if at least on constraint is not satisfied.
    Here, the function $\mathcal{J}(\balpha, \bbeta)$ is jointly strictly convex in $(\balpha, \bbeta)$.
    Hence,
    \begin{align*}
        \mathcal{J}(\balpha^t, \bbeta^t) &= \mathcal{J}\left(t(\balpha^*, \bbeta^*) + (1 - t)(\tilde{\balpha}, \bbeta)\right) \\
        &< t \mathcal{J}(\balpha^*, \bbeta^*) + (1 - t)\mathcal{J}(\tilde{\balpha}, \tilde{\bbeta}),
    \end{align*}
    for $t \in (0, 1)$ and $(\balpha^*, \bbeta^*) \neq (\tilde{\balpha}, \tilde{\bbeta})$.
    If we assume $\mathcal{J}(\balpha^*, \bbeta^*) \geq \mathcal{J}(\tilde{\balpha}, \tilde{\bbeta})$, then for $t \approx 1$, the RHS is basically $t \mathcal{J}(\balpha^*, \bbeta^*)$, while LHS is strictly smaller.
    This means
    \begin{align*}
        \mathcal{J}(\balpha^t, \bbeta^t) < \mathcal{J}(\balpha^*, \bbeta^*),
    \end{align*}
    for $t$ slightly less than $1$.
    This contradicts the minimality of $(\balpha^*, \bbeta^*) \in \Gamma^\circ$ and concludes the proof.
\end{proof}

\begin{proof}[Proof for Theorem~\ref{thm:hp-convergence-c2fa}]
    Define
    \begin{align*}
        \mathcal{R}(\balpha,\bbeta)
        \;=\;
        \mathbb{E}\!\Bigl[
        \tfrac{1}{2} \bigl\|\,Y^{\mathrm{H}} - \mathbf{Z}^{\mathrm{H}}\balpha \bigr\|^2
        \Bigr]
        \;+\;
        \mathbb{E}\!\Bigl[
        \tfrac{1}{2} \bigl\|\,Y^{\mathrm{L}} - \mathbf{Z}^{\mathrm{L}}\bbeta \bigr\|^2
        \Bigr],
    \end{align*}
    subject to the linear constraints $\alpha_j=\sum_{d=1}^{D_j}\beta_{jd}$. 
    By assumption, 
    \begin{align*}
        (\balpha^\star,\bbeta^\star) 
        \;=\; 
        \argmin_{\alpha_j=\sum_d\beta_{jd}}\;\mathcal{R}(\balpha,\bbeta).
    \end{align*}
    Meanwhile, in finite samples, the empirical counterpart is
    \begin{align*}
        \widehat{\mathcal{R}}(\balpha,\bbeta)
        \;=\;
        \underbrace{\tfrac12
        \bigl\|Y^{\mathrm{H}} - \mathbf{Z}^{\mathrm{H}}\balpha \bigr\|_{\mathbf{W}^{\mathrm{H}}}^2}_{\Lin(\balpha)}
        \;+\;
        \underbrace{\tfrac12
        \bigl\|Y^{\mathrm{L}} - \mathbf{Z}^{\mathrm{L}}\bbeta \bigr\|_{\mathbf{W}^{\mathrm{L}}}^2}_{\Lfe(\bbetad)},
    \end{align*}
    where $\|v\|_{\mathbf{W}}^2 := v^\top \mathbf{W}\,v$ for diagonal $\mathbf{W}$. 
    We also add convex regularizers 
    $\lambda_{\mathrm{H}}\,\Omega_{\mathrm{H}}(\balpha) 
    + \lambda_{\mathrm{L}}\,\Omega_{\mathrm{L}}(\bbeta)$ 
    but omit them from the basic population-risk notation for clarity. 
    Hence the C2FA solution is
    \begin{align*}
        (\widehat{\balpha},\widehat{\bbeta})
        \;=\;
        \argmin_{\alpha_j=\sum_d\beta_{jd}}\;
        \Bigl\{
        \widehat{\mathcal{R}}(\balpha,\bbeta)
        \;+\;
        \lambda_{\mathrm{H}}\,\Omega_{\mathrm{H}}(\balpha)
        \;+\;
        \lambda_{\mathrm{L}}\,\Omega_{\mathrm{L}}(\bbeta)
        \Bigr\}.
    \end{align*}

    By assumption, each pair $(\mathbf{z}^{\mathrm{H}}_n,Y^{\mathrm{H}}_n)$ is drawn i.i.d.\ with sub-Gaussian or bounded noise, so standard concentration results for random-design linear regression apply.  
    Specifically, Freedman's or Bernstein's inequality implies that with probability at least $1-\delta/2$,
    \begin{align*}
        \sup_{\theta\in\Theta}
        \Bigl|\,
        \bigl\|Y^{\mathrm{H}} - \mathbf{Z}^{\mathrm{H}}\theta \bigr\|^2_{\mathbf{W}^{\mathrm{H}}}
        \;-\;
        \mathbb{E}\!\bigl[\bigl\|Y^{\mathrm{H}} - \mathbf{Z}^{\mathrm{H}}\theta \bigr\|^2_{\mathbf{W}^{\mathrm{H}}}\bigr]
        \Bigr|
        \;\;\le\;\;
        \Gamma(N_{\mathrm{H}},J,\delta),
    \end{align*}
    where $\Gamma(\cdot)$ grows on the order of $\sqrt{(J+\log(1/\delta))\,N_{\mathrm{H}}}$.  An analogous bound holds for $(\mathbf{z}^{\mathrm{L}}_n,Y^{\mathrm{L}}_n)$ with dimension $\Dd$, also with probability at least $1-\delta/2$.  By a union bound, both events hold with probability $\ge1-\delta$.  
    Hence, for all $(\balpha,\bbeta)$ in the constraint set, we have with high probability,
    \begin{align*}
        \bigl|\,
        \widehat{\mathcal{R}}(\balpha,\bbeta)
        \;-\;
        \mathcal{R}(\balpha,\bbeta)
        \bigr|
        \;\;\le\;\;
        C_1\sqrt{\frac{(J+\Dd)+\log(1/\delta)}{\,N_{\mathrm{H}}+N_{\mathrm{L}}}},
    \end{align*}
    for some constant $C_1>0$ (absorbing weighting and separate bounds from $H/L$).
    We impose that $\mathbf{Z}^{\mathrm{H}}$ and $\mathbf{Z}^{\mathrm{L}}$ (and the combined system) satisfy a restricted eigenvalue or similar invertibility condition on the subspace defined by $\alpha_j=\sum_{d}\beta_{jd}$.  Concretely, there is a positive constant $\kappa>0$ (w.h.p.) such that for all $(\Delta_\alpha,\Delta_\beta)$ in the linear constraint subspace,
    \begin{align*}
        (\Delta_\alpha,\Delta_\beta)^\top 
        \nabla^2 \widehat{\mathcal{R}}\bigl(\balpha,\bbeta\bigr)
        (\Delta_\alpha,\Delta_\beta)
        \;\ge\;\kappa\,\|(\Delta_\alpha,\Delta_\beta)\|^2,
    \end{align*}
    provided $(\balpha,\bbeta)$ is near the solution.
    This implies strong convexity of $\widehat{\mathcal{R}}$ plus regularization, ensuring unique minimizers and stable parameter recovery.  

    Define $\Delta_\alpha := \widehat{\balpha}-\balpha^\star$ and $\Delta_\beta \coloneqq \widehat{\bbeta}-\bbeta^\star$.
    We want to show $\|(\Delta_\alpha,\Delta_\beta)\|\le O\!\Bigl(\sqrt{\tfrac{J+\Dd+\log(1/\delta)}{N_{\mathrm{H}}+N_{\mathrm{L}}}}\Bigr)$.  
    Note that $(\widehat{\balpha},\widehat{\bbeta})$ and $(\balpha^\star,\bbeta^\star)$ both satisfy $\alpha_j=\sum_d\beta_{jd}$, so $(\Delta_\alpha,\Delta_\beta)$ lies in the same affine subspace.  
    By definition of $(\widehat{\balpha},\widehat{\bbeta})$ as an empirical minimizer,
    \begin{align*}
        &\widehat{\mathcal{R}}\bigl(\widehat{\balpha},\widehat{\bbeta}\bigr)
        +
        \lambda_{\mathrm{H}}\,\Omega_{\mathrm{H}}(\widehat{\balpha})
        +
        \lambda_{\mathrm{L}}\,\Omega_{\mathrm{L}}(\widehat{\bbeta})
        \\
        &\quad\quad\quad\quad\leq
        \widehat{\mathcal{R}}(\balpha^\star,\bbeta^\star)
        +
        \lambda_{\mathrm{H}}\,\Omega_{\mathrm{H}}(\balpha^\star)
        +
        \lambda_{\mathrm{L}}\,\Omega_{\mathrm{L}}(\bbeta^\star).
    \end{align*}
    Meanwhile, $\mathcal{R}(\balpha^\star,\bbeta^\star)$ is the global minimum in the population sense.  
    Combining these with the concentration result, a standard argument in stochastic convex optimization implies
    \begin{align*}
        \mathcal{R}\bigl(\widehat{\balpha},\widehat{\bbeta}\bigr)
        -
        \mathcal{R}\bigl(\balpha^\star,\bbeta^\star\bigr)
        \leq
        C_2\,\sqrt{\frac{(J+\Dd)+\log(1/\delta)}{N_{\mathrm{H}}+N_{\mathrm{L}}}},
    \end{align*}
    for some $C_2>0$.
    By strong convexity, a gap in the population risk implies a proportional bound on $\|(\Delta_\alpha,\Delta_\beta)\|$.
    More precisely, for $\kappa$-strong convexity we get
    \begin{align*}
        \kappa\,\|(\Delta_\alpha,\Delta_\beta)\|^2 &\leq \mathcal{R}(\widehat{\balpha},\widehat{\bbeta}) - \mathcal{R}(\balpha^\star,\bbeta^\star) \\
        &\leq C_2\,\sqrt{\frac{(J+\Dd)+\log(1/\delta)}{N_{\mathrm{H}}+N_{\mathrm{L}}}}.
    \end{align*}
    Hence,
    \begin{align*}
        \bigl\|\bigl(\widehat{\balpha},\widehat{\bbeta}\bigr) 
       - \bigl(\balpha^\star,\bbeta^\star\bigr)\bigr\| 
        \leq
        \sqrt{\frac{C_2}{\kappa}}
        \;\sqrt{\frac{(J+\Dd)+\log(1/\delta)}{N_{\mathrm{H}}+N_{\mathrm{L}}}}.
    \end{align*}
    Set $C \coloneqq \sqrt{C_2/\kappa}$, and it concludes the proof.
\end{proof}

\begin{proof}[Proof for Corollary~\ref{cor:uniform-approx}]
    By Theorem~\ref{thm:hp-convergence-c2fa}, with probability at least $1-\delta/2$ we have
    \begin{align*}
        \bigl\|(\widehat{\balpha},\widehat{\bbeta}) 
         - (\balpha^\star,\bbeta^\star)\bigr\| \leq
        C_*\,\sqrt{\frac{J+\Dd + \log(1/\delta)}{\,N_{\mathrm{H}} + N_{\mathrm{L}}\,}},
    \end{align*}
    for some constant $C_*$.
    Consider any $\mathbf{z}^{\mathrm{H}} \in \mathcal{Z}^{\mathrm{H}}$.
    We write
    \begin{align*}
        &\bigl|\,e^{\mathrm{H}}(\mathbf{z}^{\mathrm{H}}) 
        - f(\phi^{\mathrm{H}}_{\mathbf{x}}(\mathbf{z}^{\mathrm{H}}))\bigr| \\
        &\quad\quad\quad\quad \leq 
        \underbrace{
        \bigl|\,\mathbf{z}^{\mathrm{H}}{}^\top \widehat{\balpha}
         - \mathbf{z}^{\mathrm{H}}{}^\top \balpha^\star\bigr|
        }_{\text{(I)}} +
        \underbrace{
        \bigl|\,\mathbf{z}^{\mathrm{H}}{}^\top \balpha^\star
         - f(\phi^{\mathrm{H}}_{\mathbf{x}}(\mathbf{z}^{\mathrm{H}}))\bigr|
        }_{\text{(II)}}.
    \end{align*}
    Term (I) is bounded by $\|\mathbf{z}^{\mathrm{H}}\|\cdot\|\widehat{\balpha}-\balpha^\star\|$, which is at most $B_{\mathrm{H}}\cdot\|\widehat{\balpha}-\balpha^\star\|$ if $\|\mathbf{z}^{\mathrm{H}}\|\le B_{\mathrm{H}}$ in $\mathcal{Z}^{\mathrm{H}}$.  
    Similarly for $\mathbf{z}^{\mathrm{L}}$.  
    Term (II) measures the approximation gap of the \emph{population-optimal} $\balpha^\star$ relative to $f$.
    One may define 
    \begin{align*}
        \epsilon^\star_{\mathrm{H}}\coloneqq \sup_{\mathbf{z}^{\mathrm{H}}\in\mathcal{Z}^{\mathrm{H}}}
        \bigl|\mathbf{z}^{\mathrm{H}}{}^\top \balpha^\star
            - f(\phi^{\mathrm{H}}_{\mathbf{x}}(\mathbf{z}^{\mathrm{H}}))\bigr|.
    \end{align*}
    Likewise define $\epsilon^\star_{\mathrm{L}}$ for the low-level side.  Because $(\balpha^\star,\bbeta^\star)$ is the \emph{best} consistent linear predictor in expectation, $\epsilon^\star_{\mathrm{H}}$ and $\epsilon^\star_{\mathrm{L}}$ capture the inherent nonlinearity or mismatch of $f$ beyond linear approximability.

    Hence,
    \begin{align*}
        \sup_{\mathbf{z}^{\mathrm{H}}\in\mathcal{Z}^{\mathrm{H}}}
        \bigl|\,
        e^{\mathrm{H}}(\mathbf{z}^{\mathrm{H}})
        - f(\phi^{\mathrm{H}}_{\mathbf{x}}(\mathbf{z}^{\mathrm{H}}))
        \bigr| \leq 
        B_{\mathrm{H}}
        \,\|\widehat{\balpha}-\balpha^\star\| + \epsilon^\star_{\mathrm{H}}.
    \end{align*}
    By a union bound, with probability at least $1-\delta$,
    \begin{align*}
        &\sup_{\mathbf{z}^{\mathrm{H}}\in\mathcal{Z}^{\mathrm{H}}}
        \Bigl|\,
        e^{\mathrm{H}}(\mathbf{z}^{\mathrm{H}}) - f(\phi^{\mathrm{H}}_{\mathbf{x}}(\mathbf{z}^{\mathrm{H}}))
        \Bigr| \\
        &\quad\quad\quad \leq 
        B_{\mathrm{H}}\cdot
        C_*
        \sqrt{\frac{J+\Dd + \log(1/\delta)}{\,N_{\mathrm{H}} + N_{\mathrm{L}}\,}}
        \;+\;\epsilon^\star_{\mathrm{H}},
    \end{align*}
    and similarly for $e^{\mathrm{L}}$.  
    Add the two supremum gaps to get:
    \begin{align*}
        &\sup_{\mathbf{z}^{\mathrm{H}}\in\mathcal{Z}^{\mathrm{H}}}
        \Bigl|\,
        f(\phi^{\mathrm{H}}_{\mathbf{x}}(\mathbf{z}^{\mathrm{H}}))
        - e^{\mathrm{H}}(\mathbf{z}^{\mathrm{H}})
        \Bigr| +
        \sup_{\mathbf{z}^{\mathrm{L}}\in\mathcal{Z}^{\mathrm{L}}}
        \Bigl|\,
        f(\phi^{\mathrm{L}}_{\mathbf{x}}(\mathbf{z}^{\mathrm{L}}))
        - e^{\mathrm{L}}(\mathbf{z}^{\mathrm{L}})
        \Bigr| \\
        &\quad\quad\quad \leq 
        \widetilde{C}
        \sqrt{\frac{J+\Dd + \log(1/\delta)}{N_{\mathrm{H}} + N_{\mathrm{L}}}} + 
        (\epsilon^\star_{\mathrm{H}} + \epsilon^\star_{\mathrm{L}}).
    \end{align*}
    We can absorb $\epsilon^\star_{\mathrm{H}} + \epsilon^\star_{\mathrm{L}}$ into a single constant if desired, because it is problem-dependent. Hence the final bound holds with probability at least $1-\delta$.  
\end{proof}

\section{Limitations and Broader Impacts}\label{sec:limitations}
A possible limitation of the proposed method is that the quality of the HiFAs and LoFAs may be worse in cases where the consistency property is inherently not satisfied.
For example, they may happen when the HiFAs and LoFAs are estimated with the combination of different approaches, such as MILLI and LIME, and when the behaviors of the black-box model vary significantly between perturbed inputs that high- and low-level features are partially masked.
To detect such an undesirable situation early, monitoring the losses of the surrogate models, $\Lin$ in \bref{eq:proposed:HiFA} and $\Lfe$ in \bref{eq:proposed:LoFA}, is effective because they are likely to be worse even if the objective~\bref{eq:proposed:objective1} is minimized. 

Our work contributes to improving the transparency of black-box models.
However, it should be noted that high-quality feature attributions may give hints about stealing the information that the model's providers want to hide, such as the training data and the model's decision-making process.
To prevent such risks, it is essential to establish guidelines that ensure that the feature attributions are not used for malicious purposes.

\subsection*{Acknowledgments}
This work was supported by JSPS KAKENHI Grant Number 22K17953.

\bibliographystyle{named}
\bibliography{references}

\begin{thebibliography}{}

\bibitem[\protect\citeauthoryear{Boyd \bgroup \em et al.\egroup }{2011}]{Boyd_undated-bc}
Stephen Boyd, Neal Parikh, Eric Chu, Borja Peleato, and Jonathan Eckstein.
\newblock {\em {Distributed Optimization and Statistical Learning via the Alternating Direction Method of Multipliers}}.
\newblock Now Foundations and Trends, 2011.

\bibitem[\protect\citeauthoryear{Cheplygina \bgroup \em et al.\egroup }{2019}]{Cheplygina2018-wj}
Veronika Cheplygina, Marleen {de Bruijne}, and Josien~P.W. Pluim.
\newblock {Not-so-supervised: A survey of semi-supervised, multi-instance, and transfer learning in medical image analysis}.
\newblock {\em Medical Image Analysis}, 54:280--296, 2019.

\bibitem[\protect\citeauthoryear{Danilevsky \bgroup \em et al.\egroup }{2020}]{Danilevsky2020-sy}
Marina Danilevsky, Kun Qian, Ranit Aharonov, Yannis Katsis, Ban Kawas, and Prithviraj Sen.
\newblock {A Survey of the State of Explainable AI for Natural Language Processing}.
\newblock In {\em {Proceedings of the Conference of the Asia-Pacific Chapter of the Association for Computational Linguistics and the International Joint Conference on Natural Language Processing}}, pages 447--459, 2020.

\bibitem[\protect\citeauthoryear{Dara \bgroup \em et al.\egroup }{2020}]{Dara2020-ml}
Sriharsha Dara, C~Ravindranath Chowdary, and Chintoo Kumar.
\newblock {A survey on group recommender systems}.
\newblock {\em {Journal of Intelligent Information Systems}}, 54(2):271--295, 2020.

\bibitem[\protect\citeauthoryear{Devlin \bgroup \em et al.\egroup }{2018}]{DBLP:journals/corr/abs-1810-04805}
Jacob Devlin, Ming{-}Wei Chang, Kenton Lee, and Kristina Toutanova.
\newblock {BERT:} pre-training of deep bidirectional transformers for language understanding.
\newblock {\em arXiv preprint arXiv:1810.04805}, 2018.

\bibitem[\protect\citeauthoryear{Do{\v s}ilovi{\'c} \bgroup \em et al.\egroup }{2018}]{Dosilovic2018-uh}
Filip~Karlo Do{\v s}ilovi{\'c}, Mario Br{\v c}i{\'c}, and Nikica Hlupi{\'c}.
\newblock {Explainable artificial intelligence: A survey}.
\newblock In {\em {Proceedings of the International Convention on Information and Communication Technology, Electronics and Microelectronics}}, pages 210--215, 2018.

\bibitem[\protect\citeauthoryear{Early \bgroup \em et al.\egroup }{2022}]{Early2022-oc}
Joseph Early, Christine Evers, and Sarvapali Ramchurn.
\newblock In {\em {Proceedings of the International Conference on Learning Representations}}, 2022.

\bibitem[\protect\citeauthoryear{Everingham \bgroup \em et al.\egroup }{2015}]{Everingham15}
M.~Everingham, S.~M.~A. Eslami, L.~Van~Gool, C.~K.~I. Williams, J.~Winn, and A.~Zisserman.
\newblock {The Pascal Visual Object Classes Challenge: A Retrospective}.
\newblock {\em {International Journal of Computer Vision}}, 111(1):98--136, 2015.

\bibitem[\protect\citeauthoryear{fabriceyhc (Hugging~Face)}{2022}]{Harel-Canada_undated-qi}
fabriceyhc (Hugging~Face).
\newblock {fabriceyhc/bert-base-uncased-amazon\_polarity}, 2022.
\newblock Accessed: 2024-2-2.

\bibitem[\protect\citeauthoryear{Guo \bgroup \em et al.\egroup }{2021}]{Guo2021-vp}
Yulan Guo, Hanyun Wang, Qingyong Hu, Hao Liu, Li~Liu, and Mohammed Bennamoun.
\newblock {Deep Learning for 3D Point Clouds: A Survey}.
\newblock {\em {IEEE Transactions on Pattern Analysis and Machine Intelligence}}, 43(12):4338--4364, 2021.

\bibitem[\protect\citeauthoryear{He \bgroup \em et al.\egroup }{2016}]{He2016-kb}
Kaiming He, Xiangyu Zhang, Shaoqing Ren, and Jian Sun.
\newblock {Deep residual learning for image recognition}.
\newblock In {\em {Proceedings of the IEEE/CVF Conference on Computer Vision and Pattern Recognition}}, pages 770--778, 2016.

\bibitem[\protect\citeauthoryear{Ilse \bgroup \em et al.\egroup }{2018}]{Ilse2018-be}
Maximilian Ilse, Jakub Tomczak, and Max Welling.
\newblock {Attention-based Deep Multiple Instance Learning}.
\newblock In {\em {Proceedings of the International Conference on Machine Learning}}, volume~80, pages 2127--2136, 2018.

\bibitem[\protect\citeauthoryear{Jain \bgroup \em et al.\egroup }{2022}]{Jain2022-oe}
Saachi Jain, Hadi Salman, Eric Wong, Pengchuan Zhang, Vibhav Vineet, Sai Vemprala, and Aleksander Madry.
\newblock {Missingness Bias in Model Debugging}.
\newblock In {\em {Proceedings of the International Conference on Learning Representations}}, 2022.

\bibitem[\protect\citeauthoryear{Javed \bgroup \em et al.\egroup }{2022}]{Javed2022-sm}
Syed~Ashar Javed, Dinkar Juyal, Harshith Padigela, Amaro Taylor-Weiner, Limin Yu, and Aaditya Prakash.
\newblock {Additive MIL: intrinsically interpretable multiple instance learning for pathology}.
\newblock {\em {Advances in Neural Information Processing Systems}}, pages 20689--20702, 2022.

\bibitem[\protect\citeauthoryear{Kimura \bgroup \em et al.\egroup }{2024}]{Kimura2024-wq}
Masanari Kimura, Ryotaro Shimizu, Yuki Hirakawa, Ryosuke Goto, and Yuki Saito.
\newblock {On permutation-invariant neural networks}.
\newblock {\em arXiv preprint arXiv:2403.17410}, 2024.

\bibitem[\protect\citeauthoryear{Kingma and Ba}{2015}]{Kingma2014-qn}
Diederik~P Kingma and Jimmy Ba.
\newblock {Adam: A method for stochastic optimization}.
\newblock In {\em {Proceedings of the International Conference on Learning Representations}}, 2015.

\bibitem[\protect\citeauthoryear{Lundberg and Lee}{2017}]{Lundberg2017-ii}
Scott~M Lundberg and Su-In Lee.
\newblock {A Unified Approach to Interpreting Model Predictions}.
\newblock {\em {Advances in Neural Information Processing Systems}}, pages 4765--4774, 2017.

\bibitem[\protect\citeauthoryear{Mosca \bgroup \em et al.\egroup }{2022}]{Mosca2022-ki}
Edoardo Mosca, Defne Demirt{\"u}rk, Luca M{\"u}lln, Fabio Raffagnato, and Georg Groh.
\newblock Grammar{SHAP}: An efficient model-agnostic and structure-aware {NLP} explainer.
\newblock In {\em {Proceedings of the ACL Workshop on Learning with Natural Language Supervision}}, pages 10--16, 2022.

\bibitem[\protect\citeauthoryear{Petsiuk \bgroup \em et al.\egroup }{2018}]{Petsiuk2018-uj}
Vitali Petsiuk, Abir Das, and Kate Saenko.
\newblock {RISE: Randomized Input Sampling for Explanation of Black-box Models}.
\newblock In {\em {Proceedings of the British Machine Vision Conference}}, 2018.

\bibitem[\protect\citeauthoryear{Plumb \bgroup \em et al.\egroup }{2018}]{Plumb2018-ac}
Gregory Plumb, Denali Molitor, and Ameet~S Talwalkar.
\newblock {Model Agnostic Supervised Local Explanations}.
\newblock {\em {Advances in Neural Information Processing Systems}}, pages 4768--4777, 2018.

\bibitem[\protect\citeauthoryear{Ribeiro \bgroup \em et al.\egroup }{2016}]{Ribeiro2016-nj}
Marco~Tulio Ribeiro, Sameer Singh, and Carlos Guestrin.
\newblock {"Why Should I Trust You?" Explaining the Predictions of Any Classifier}.
\newblock In {\em Proceedings of the ACM SIGKDD International Conference on Knowledge Discovery and Data Mining}, pages 1135--1144, 2016.

\bibitem[\protect\citeauthoryear{Ribeiro \bgroup \em et al.\egroup }{2018}]{Ribeiro2018-ws}
Marco~Tulio Ribeiro, Sameer Singh, and Carlos Guestrin.
\newblock {Anchors: High-Precision Model-Agnostic Explanations}.
\newblock In {\em {Proceedings of the AAAI Conference on Artificial Intelligence}}, volume 32(1), 2018.

\bibitem[\protect\citeauthoryear{Rychener \bgroup \em et al.\egroup }{2023}]{Rychener2020-gu}
Yves Rychener, Xavier Renard, Djam{\'e} Seddah, Pascal Frossard, and Marcin Detyniecki.
\newblock {On the Granularity of Explanations in Model Agnostic NLP Interpretability}.
\newblock In {\em {Proceedings of the Machine Learning and Principles and Practice of Knowledge Discovery in Databases}}, pages 498--512, 2023.

\bibitem[\protect\citeauthoryear{Saeed and Omlin}{2023}]{Saeed2023-pn}
Waddah Saeed and Christian Omlin.
\newblock {Explainable AI (XAI): A systematic meta-survey of current challenges and future opportunities}.
\newblock {\em {Knowledge-Based Systems}}, 263:110273, 2023.

\bibitem[\protect\citeauthoryear{Sampaio and Cordeiro}{2023}]{Sampaio2023-ji}
Vicente Sampaio and Filipe~R. Cordeiro.
\newblock {Improving Mass Detection in Mammography Images: A Study of Weakly Supervised Learning and Class Activation Map Methods}.
\newblock In {\em {Proceedings of the Conference on Graphics, Patterns and Images}}, pages 139--144, 2023.

\bibitem[\protect\citeauthoryear{{shap (GitHub)}}{2024}]{noauthor_undated-ca}
{shap (GitHub)}.
\newblock {shap: A game theoretic approach to explain the output of any machine learning model}, 2024.
\newblock Accessed: 2024-2-2.

\bibitem[\protect\citeauthoryear{Tan and Kotthaus}{2022}]{Tan2022-ki}
Hanxiao Tan and Helena Kotthaus.
\newblock {Surrogate model-based explainability methods for point cloud NNs}.
\newblock In {\em {Proceedings of the IEEE/CVF Winter Conference on Applications of Computer Vision}}, pages 2927--2936, 2022.

\bibitem[\protect\citeauthoryear{van~der Velden \bgroup \em et al.\egroup }{2022}]{Van_der_Velden2022-mk}
Bas H~M van~der Velden, Hugo~J Kuijf, Kenneth G~A Gilhuijs, and Max~A Viergever.
\newblock {Explainable artificial intelligence (XAI) in deep learning-based medical image analysis}.
\newblock {\em {Medical image analysis}}, 79:102470, 2022.

\bibitem[\protect\citeauthoryear{Vedaldi and Soatto}{2008}]{Vedaldi2008-us}
Andrea Vedaldi and Stefano Soatto.
\newblock {Quick Shift and Kernel Methods for Mode Seeking}.
\newblock In {\em {Proceedings of the European Conference on Computer Vision}}, pages 705--718, 2008.

\bibitem[\protect\citeauthoryear{Zafar \bgroup \em et al.\egroup }{2021}]{Zafar2021-aq}
Muhammad~Bilal Zafar, Philipp Schmidt, Michele Donini, C{\'e}dric Archambeau, Felix Biessmann, Sanjiv~Ranjan Das, and Krishnaram Kenthapadi.
\newblock {More Than Words: Towards Better Quality Interpretations of Text Classifiers}.
\newblock {\em arXiv preprint arXiv:2112.12444}, 2021.

\bibitem[\protect\citeauthoryear{Zaheer \bgroup \em et al.\egroup }{2017}]{Zaheer2017-in}
Manzil Zaheer, Satwik Kottur, Siamak Ravanbakhsh, Barnabas Poczos, Ruslan~R Salakhutdinov, and Alexander~J Smola.
\newblock {Deep Sets}.
\newblock {\em {Advances in Neural Information Processing Systems}}, pages 3391--3401, 2017.

\bibitem[\protect\citeauthoryear{Zhang \bgroup \em et al.\egroup }{2015}]{Zhang2015-sm}
Xiang Zhang, Junbo Zhao, and Yann LeCun.
\newblock {Character-level convolutional networks for text classification}.
\newblock {\em {Advances in Neural Information Processing Systems}}, 28:649--657, 2015.

\end{thebibliography}

\end{document}